%% file: main.tex
\documentclass[10pt,twocolumn,letterpaper]{article}

\usepackage[pagenumbers]{cvpr} %

\input{preamble}

\definecolor{cvprblue}{rgb}{0.21,0.49,0.74}
\usepackage[pagebackref,breaklinks,colorlinks,allcolors=cvprblue]{hyperref}
\usepackage{graphicx}
\usepackage{cuted}
\usepackage{caption}

\title{M\textsuperscript{2}-WAM: Efficient Hierarchical Memory Across Modalities and Timescales for World-Action Models}

\author{Xiaopeng Zhang$^{1,2}$ \quad Weiqi Yuan$^{1}$ \quad Xin Zhu$^{2,\dagger,\ddagger}$ \quad Jiaxi Huang$^{2}$ \quad Yueqi Hu$^{2}$ \\
Yin Zhang$^{2}$ \quad Jian Yao$^{2}$ \quad Yanjie Li$^{1,\dagger}$ \\[4pt]
$^{1}$Harbin Institute of Technology, Shenzhen \quad $^{2}$Robotics Foundation Model Team, XPENG
}

\begin{document}
\maketitle
\begingroup
\renewcommand{\thefootnote}{}
\begin{NoHyper}
\footnotetext[0]{$^{\dagger}$Corresponding authors. \quad $^{\ddagger}$Project leader.}
\end{NoHyper}
\endgroup
\input{sec/0_abstract}    
\input{sec/1_intro}
\begin{figure*}[t]
    \centering
    \includegraphics[width=\textwidth]{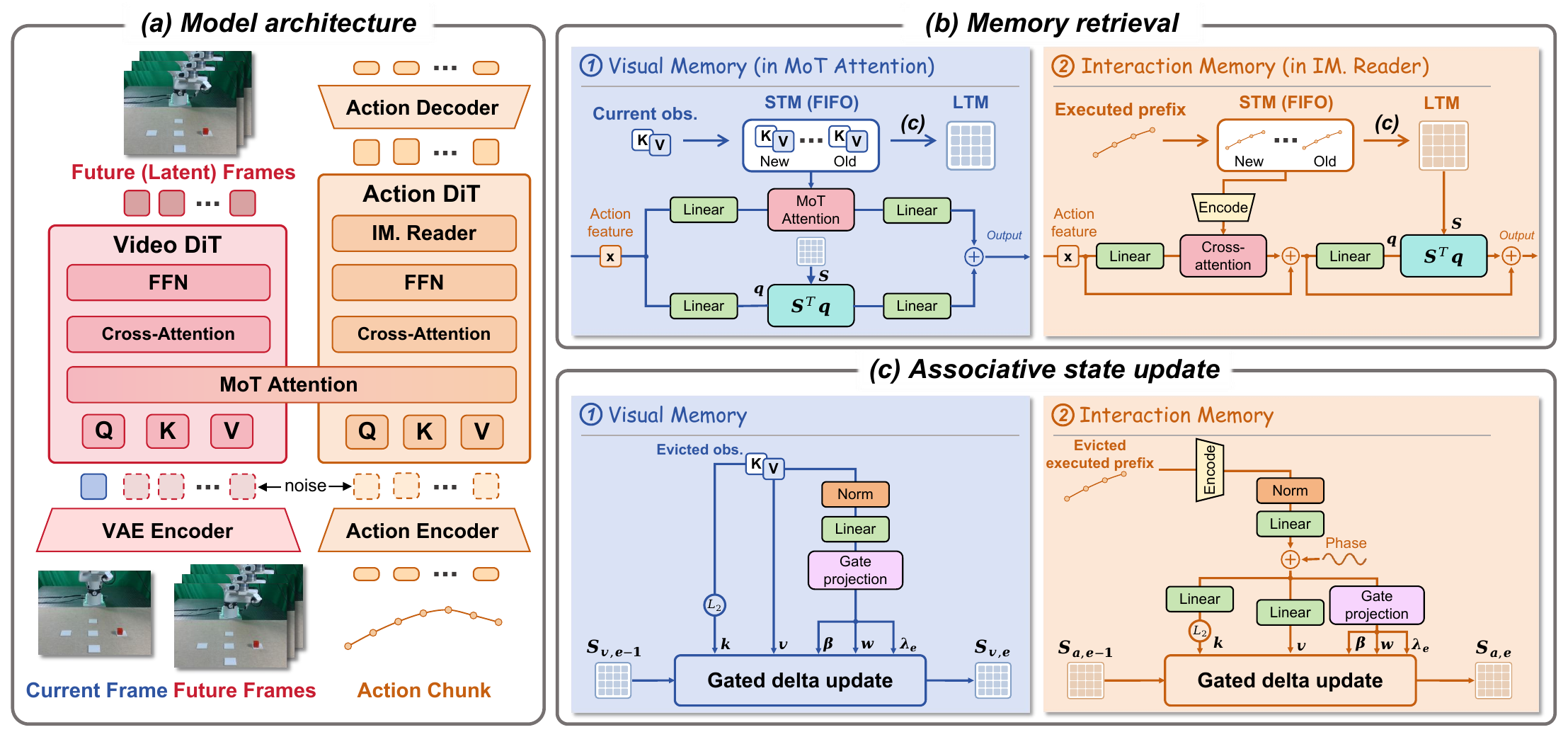}
    \caption{\textbf{Overview of the M\textsuperscript{2}-WAM architecture.}
    \textbf{(a)} Video and action diffusion transformers form a mixture-of-transformers backbone; future-video prediction is training-only.
    \textbf{(b)} Action features retrieve visual memory within MoT attention and interaction memory (IM) through a reader. Both combine short-term memory (STM) attention with associative long-term memory (LTM) readouts.
    \textbf{(c)} Modality-specific writers consolidate evicted visual key-value entries and executed prefixes into fixed-size LTM states using a common gated delta update rule.
    Blue and orange indicate visual and interaction memory, respectively.
    Some normalization and scaling operations are omitted.}
    \label{fig:architecture}
\end{figure*}
\input{sec/2_related_work}
\input{sec/3_method}
\input{sec/4_experiments}
\input{sec/5_conclusion}

\clearpage
{
    \small
    \bibliographystyle{ieeenat_fullname}
    \bibliography{main}
}

\input{sec/X_suppl}

\end{document}

%% file: preamble.tex
\usepackage{multirow}
\usepackage{algorithm}
\usepackage{algpseudocode}

\definecolor{ourslavender}{HTML}{F1EDF8}
\definecolor{fastwamblue}{HTML}{E5EFFA}
\makeatletter
\newcommand{\resultrow}[2]{\noalign{\begingroup
  \color{#1}\hrule
  \if\relax\detokenize{#2}\relax\else width #2\fi
  height \dimexpr\ht\@arstrutbox+\dp\@arstrutbox\relax
  \kern-\dimexpr\ht\@arstrutbox+\dp\@arstrutbox\relax
\endgroup}}
\newcommand{\oursrow}[1]{\resultrow{ourslavender}{#1}}
\newcommand{\fastwamrow}[1]{\resultrow{fastwamblue}{#1}}
\makeatother
\newcommand{\avgdelta}[1]{\,{\scriptsize$(#1)$}}

%% file: sec/0_abstract.tex
\begin{strip}
    \centering
    \includegraphics[width=\textwidth]{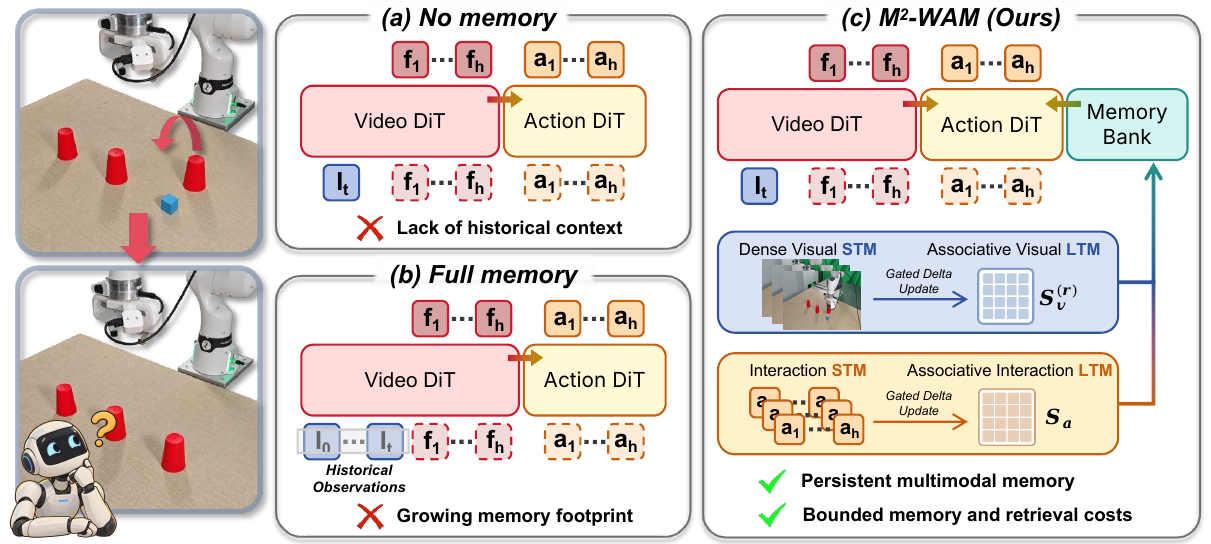}
    \captionsetup{hypcap=false}
    \captionof{figure}{
        \textbf{Motivation and overview of M\textsuperscript{2}-WAM.} (a) Memoryless WAMs lack historical context. (b) Full-history memory incurs growing storage and retrieval costs. (c) M\textsuperscript{2}-WAM organizes visual and interaction histories into short-term memory (STM) and fixed-size associative long-term memory (LTM), enabling persistent history conditioning with bounded online memory costs.
    }
    \label{fig:overview}
    \captionsetup{hypcap=true}
\end{strip}
\begin{abstract}
Long-horizon robotic manipulation requires historical context, yet retaining extensive visual histories incurs growing storage and retrieval costs, while aggressive compression risks losing fine-grained information needed for subsequent decisions.
Moreover, executed action histories capture intermediate behaviors that sparse visual observations may miss.
Preserving this complementary context efficiently calls for modality-specific memory mechanisms that balance recent detail with long-term retention.
To this end, we introduce \textbf{M\textsuperscript{2}-WAM}, a world-action model equipped with hierarchical visual and interaction memories.
M\textsuperscript{2}-WAM preserves recent visual observations in dense short-term memory while consolidating entries evicted from this window into fixed-size associative long-term memory through gated delta updates.
Within the same short- and long-term hierarchy, interaction memory encodes executed action prefixes into compact temporal representations, complementing visual observations with behavioral context.
These complementary memories are integrated into world-action modeling through layer-wise retrieval, enabling memory-augmented action generation with a bounded online memory footprint independent of episode length.
On the memory-dependent simulation benchmarks MIKASA-Robo and RMBench, M\textsuperscript{2}-WAM achieves +53.2\% and +61.2\% absolute improvements in average success rates over Fast-WAM, respectively.
These gains extend to real-world manipulation, where M\textsuperscript{2}-WAM achieves a +29.3\% absolute improvement over LingBot-VA in average success across three tasks.
These results suggest that M\textsuperscript{2}-WAM offers a promising approach to efficient memory-augmented world-action modeling.
\end{abstract}

%% file: sec/1_intro.tex
\section{Introduction}
\label{sec:intro}

Long-horizon robotic manipulation often requires reasoning beyond the current observation \cite{zheng2025tracevla,xiao2026avavla,lin2026hifvla}. For example, in a shell game task, a robot must track a target object as it is occluded and moved beneath visually identical cups, relying on historical observations to identify its location \cite{cherepanov2026memory,guo2026chameleon}. Such tasks are inherently non-Markovian with respect to instantaneous observations, requiring policies to reason over past observations and interactions \cite{kaelbling1998pomdp}. Recent world-action models (WAMs) have demonstrated promising capabilities by jointly modeling visual dynamics and robot actions \cite{ye2026worldactionmodelszeroshot,lingbot-va2026,yuan2026fastwam}. However, many WAMs still rely on current observations or limited contexts, making it difficult to exploit information from earlier stages of execution.

Historical context must support ongoing decisions while preserving
information from earlier task stages. Recent observations provide
spatial detail, whereas earlier observations may
contain cues that remain relevant after they leave view.
As shown in Figure~\ref{fig:overview}(a,b), a memoryless WAM lacks
access to earlier observations, whereas retaining full visual history
incurs storage and retrieval costs that grow with episode length.
A sliding window preserves recent detail but discards older observations
\cite{wang2026nativemem,yang2026eventvla}.
Alternatively, compact or recurrent memories reduce storage costs,
but compression can limit access to fine-grained historical information
\cite{yang2026memorywam,zhao2026dram}.
We argue for allocating memory capacity across timescales:
retaining recent observations explicitly and consolidating older
entries into a compact, persistent state. This design preserves
recent detail without compressing all observations and
extends historical coverage beyond a fixed window, drawing inspiration
from short- and long-term memory \cite{atkinson1968human} and
complementary learning systems \cite{mcClelland1995why}.

However, retaining history across timescales is only part of the
challenge: memory must also capture complementary information
from observations and executed actions. Visual observations
describe scene configurations and changes, but may omit behavior
between sampled frames.
Denser sampling increases visual encoding and storage costs
for the same duration.
Executed action prefixes complement these observations by
recording the intervening actions, providing compact, temporally
denser behavioral context
\cite{zhou2026action,zhao2026remember,pala2026membodied}.
This motivates modality-aware memory that preserves
both scene observations and executed actions.
We therefore adopt visual and interaction memories that preserve
spatial details and temporal action patterns, respectively,
within a common short- and long-term hierarchy.

Building on these considerations, we propose \textbf{M\textsuperscript{2}-WAM}, which augments a mixture-of-transformers (MoT) architecture with complementary visual and interaction memories across timescales, as illustrated in Figure~\ref{fig:overview}(c). The visual memory preserves an initial observation anchor and recent observations as dense spatial key-value (KV) representations, while consolidating older information into fixed-size, layer-wise associative states through gated delta updates \cite{yang2025gated,hatamizadeh2026gated}. To accommodate robot image sequences, we apply temporal decay once per observation according to elapsed time, rather than repeatedly across its spatial tokens. We further replace channel-wise erasure with a scalar erase gate per token and head to improve numerical stability. In parallel, the interaction memory encodes executed action prefixes into compact temporal representations and maintains a shared associative state. These memories are integrated into world-action modeling through layer-wise, query-conditioned retrieval. By combining bounded short-term memories with fixed-size associative states, M\textsuperscript{2}-WAM enables persistent history conditioning with online memory storage and retrieval costs independent of episode length.

Our main contributions are as follows:
1) We propose M\textsuperscript{2}-WAM, a world-action model equipped with modality-specialized hierarchical memory that jointly preserves visual observations and executed interactions across short and long timescales.
2) We develop modality-specific memory representations and associative consolidation mechanisms, together with complementary retrieval pathways that incorporate historical information into world-action computation with bounded online memory and retrieval costs.
3) We conduct extensive experiments on MIKASA-Robo
\cite{cherepanov2026memory}, RMBench \cite{chen2026rmbench},
RoboTwin~2.0 \cite{chen2025robotwin}, and three real-world tasks,
complemented by ablation studies and qualitative and quantitative
analyses. Results demonstrate improved performance on memory-dependent
tasks with bounded online memory and retrieval costs,
supporting the effectiveness and efficiency of our method.

%% file: sec/2_related_work.tex
\section{Related Work}
\label{sec:related_work}

\subsection{World-Action Models}
Recent robot policies learn compact state representations \cite{liu2026stamo}, couple visual prediction with policy learning \cite{tian2025seer,hu2025vpp,cheang2024gr2generativevideolanguageactionmodel}, or plan with predicted future states \cite{du2023learning,zhao2025cotvla,zhang2026foreact,huang2026pointworld}. World-action models connect visual dynamics prediction with action generation through unified generative models or coupled video–action architectures \cite{zhu2025uwm,cen2025WorldVLA,bi2025motusunifiedlatentaction,kim2025cosmospolicy}. DreamZero and LingBot-VA demonstrate the potential of video-based world modeling for generalizable robot control \cite{ye2026worldactionmodelszeroshot,lingbot-va2026}. Recent studies further investigate video-action coupling and its role in policy learning and task transfer \cite{ma2026dit4dit,wang2026patchwam,cen2026worldbridgebridgingvideoaction,OneWM}. Inference efficiency is also a practical concern for robot policies \cite{zhang2025molevla}. Fast-WAM retains video co-training but skips future prediction at test time, whereas Faster-WAM preserves future conditioning through selective video-action interactions \cite{yuan2026fastwam,zhao2026faster}. Building on Fast-WAM, we investigate how visual and executed-action histories can support action generation under a bounded online memory budget.

\subsection{Memory-Augmented Robotic Policies}
Memory-augmented robotic policies address partial observability through explicit memory banks, compressed visual histories, and selective retrieval \cite{yin2026simplememvla,wang2026nativemem,gao2026gatedmemorypolicy,cen2026worldbridgebridgingvideoaction,li2026trmvla,wang2026dual}. MemoryVLA maintains perceptual and cognitive memory streams, retrieving historical context with current tokens and consolidating redundant entries \cite{shi2025memoryvla}. MemoryVLA++ combines this memory with imagined future latents to condition action generation \cite{shi2026memoryvla++}. BridgeVLA++ pairs temporal memory of historical keyframes with spatial memory of the initial scene geometry, supporting task-progress reasoning and localization under occlusion \cite{li2026bridgevlaplus}. MemoryWAM combines recent observations, anchor frames, and historical gist tokens \cite{yang2026memorywam}, whereas HiMem-WAM stores compact task states at predicted skill boundaries \cite{sun2026himemwam}. To retain history without accumulating tokens, recurrent approaches maintain fixed-size associative states using gated delta updates \cite{yang2025gated,hatamizadeh2026gated}. DRAM jointly incorporates tokens within each observation \cite{zhao2026dram}, while MemBodied combines action summaries with their observed visual consequences \cite{pala2026membodied}. The latter also highlights the relevance of behavioral history, explored through joint visual-action representations in AEM and executed-action memory in ActMem-VLA \cite{zhou2026action,zhao2026remember}. M\textsuperscript{2}-WAM organizes visual observations and executed action prefixes into separate pathways, each combining explicit short-term storage with associative long-term consolidation. Dense visual KV features and compact temporal action representations are retained and retrieved through modality-specific structures, while entries evicted from each short-term window update its corresponding fixed-size associative state.

%% file: sec/3_method.tex
\section{Method}
\label{sec:method}

\subsection{Problem Formulation and Architecture}
We consider long-horizon robotic manipulation under partial observability,
where action generation may depend on information beyond the current observation.
At each replanning step $t$, the robot receives an observation
$o_t=(\mathcal{I}_t,\mathbf{s}_t)$, comprising multi-view images and
proprioceptive state, and a task instruction $\ell$. Given a bounded
memory bank $\mathcal{M}_t$ representing past observations and executed actions, the
policy predicts an action chunk $\mathbf{A}_t\in\mathbb{R}^{H\times d_a}$:
\begin{equation}
    \pi_\theta(\mathbf{A}_t\mid o_t,\ell,\mathcal{M}_t).
\end{equation}
Here, $H$ is the prediction horizon and $d_a$ is the action dimension.
The online memory satisfies $\operatorname{Size}(\mathcal{M}_t)\le B$,
where $B$ is a fixed budget independent of episode length.

As illustrated in Figure~\ref{fig:architecture},
M\textsuperscript{2}-WAM builds on Fast-WAM's MoT backbone, comprising
a video Diffusion Transformer (DiT) \cite{peebles2023dit} and an action DiT. During inference, the video DiT encodes the current
observation into layer-wise visual KV representations that condition
iterative action denoising in the action DiT, without generating future video.

We augment this architecture with complementary visual and interaction
memories, each combining bounded short-term memory (STM) with fixed-size
associative long-term memory (LTM). The action DiT directly retrieves
historical context from both memory streams during denoising.

\subsection{Short-Term Visual and Interaction Memory}
Visual and interaction STM preserve recent visual context and executed
action prefixes in separate bounded first-in, first-out (FIFO) queues. Let $\mathcal{W}_t$ denote the ordered indices
of retained historical entries before replanning step $t$, with
$|\mathcal{W}_t|\le N$. Each entry pairs a sampled observation with the
action prefix executed after that observation.
The effect of FIFO length is examined in Appendix~\ref{app:fifo_length}.

\noindent\textbf{Visual memory.} For each observation $i$, we cache the dense spatial KV tensors
$(\mathbf{K}_i^{(r)},\mathbf{V}_i^{(r)})$ produced at every
layer $r$ of the video DiT. Each observation is encoded without reading historical memory,
and its cached representations retain the spatial token layout
without additional pooling. Visual STM maintains these layer-wise KV
entries in chronological order. In addition, the KV representations of
the episode's first observation form a permanent visual anchor, retained
outside the FIFO without consuming its $N$ slots.

\noindent\textbf{Interaction memory.} Interaction STM stores normalized executed action
prefixes $\mathbf{A}_i^{\mathrm{exec}}$ in a matching FIFO. For retrieval,
a temporal encoder applies one-dimensional convolutions
and adaptive average pooling, followed by a linear projection and layer
normalization:
\begin{equation}
    \mathbf{Z}_i = E_\psi(\mathbf{A}_i^{\mathrm{exec}})
    \in \mathbb{R}^{P\times d_m},
\end{equation}
where $P$ is the number of tokens per executed prefix and $d_m$ is their
feature dimension. The resulting tokens
summarize successive temporal regions and provide
an explicit representation of recent robot behavior for retrieval.

\subsection{Associative Long-Term Memory}
\label{sec:associative_ltm}
When advancing a full FIFO, we consolidate its oldest entry into a
fixed-size associative state before eviction. Visual LTM maintains
an independent multi-head state at each layer of the video DiT, while interaction
LTM maintains one multi-head state shared across layers of the action DiT.
The writers are independently parameterized and follow a common update
rule. Below, we describe one head, omitting modality and layer indices.
Its state $\mathbf{S}\in\mathbb{R}^{d_k\times d_v}$, where $d_k$ and $d_v$
are the per-head key and value dimensions, is initialized to zero at the
start of each episode.

\noindent\textbf{Writing representations.}
For visual writes, we reuse the video DiT's native keys before rotary
position embeddings (RoPE) \cite{su2024roformer}, together with its native values.
The writer applies per-head $\ell_2$ normalization to the keys and predicts
its gates from normalized and projected value features. For interaction
writes, the action content tokens $\mathbf{Z}_i$ are normalized and
projected into writer features, augmented with sinusoidal within-chunk
phase encodings, and mapped to memory keys and values.
The keys are also $\ell_2$-normalized.
Gate projection in Figure~\ref{fig:architecture}(c) predicts erase
and write gates and decay rates.

\noindent\textbf{Gated delta update.}
We build on Gated DeltaNet-2 \cite{hatamizadeh2026gated}, adapting its erase
granularity and temporal decay to WAM memory,
as illustrated in Figure~\ref{fig:architecture}(c).
We predict one scalar erase gate per token and head, shared across
key channels, while retaining independent channel-wise writing.
Scalar erasure avoids the numerical instability we observe with channel-wise erasure during training.

Spatial tokens from the same observation share a timestamp, so we apply
temporal decay once per evicted entry rather than once per spatial token.
The same entry-level decay applies to interaction memory.
For each evicted entry $e$, we first decay the memory state:
\begin{equation}
\begin{aligned}
    \mathbf{D}_e
        &= \operatorname{diag}
           \left(\exp(-\Delta_e\boldsymbol{\lambda}_e)\right), \\
    \mathbf{S}_{e,0}
        &= \mathbf{D}_e\mathbf{S}_{e-1}.
\end{aligned}
\end{equation}
Here, $\boldsymbol{\lambda}_e\in\mathbb{R}_{>0}^{d_k}$ contains
input-dependent decay rates predicted from writer features averaged over
the entry's tokens. $\Delta_e$ is the source-frame interval since the preceding
written entry, normalized by the memory sampling stride and set to one
for the first write. Interaction entries use their paired observations'
source-frame indices, so decay in both streams follows elapsed source time.

We then process the $J_e$ tokens in entry $e$ sequentially in flattened spatial
or temporal order, indexed by $j=1,\ldots,J_e$. Let
$\mathbf{k}_{e,j}\in\mathbb{R}^{d_k}$ and
$\mathbf{v}_{e,j}\in\mathbb{R}^{d_v}$ denote the token's key and value,
respectively, with $\|\mathbf{k}_{e,j}\|_2=1$. The update is
\begin{equation}
\begin{aligned}
    \mathbf{S}_{e,j}
        &= \left(\mathbf{I}
           -\beta_{e,j}\mathbf{k}_{e,j}\mathbf{k}_{e,j}^{\top}\right)
           \mathbf{S}_{e,j-1} \\
        &\quad + \mathbf{k}_{e,j}
           \left(\mathbf{w}_{e,j}\odot\mathbf{v}_{e,j}\right)^{\top}.
\end{aligned}
\end{equation}
Here, $\mathbf{I}$ is the $d_k\times d_k$ identity matrix and $\odot$
denotes element-wise multiplication. The memory predicts the value
associated with the incoming key as
$\mathbf{S}_{e,j-1}^{\top}\mathbf{k}_{e,j}$. The update applies a
rank-one correction along $\mathbf{k}_{e,j}$, subtracting this prediction
under the scalar erase gate $\beta_{e,j}\in(0,1)$ and inserting the
incoming value under the independent channel-wise write gate
$\mathbf{w}_{e,j}\in(0,1)^{d_v}$.
Both gates are predicted by separate sigmoid projections of the token's
writer feature. After all $J_e$ tokens, the updated state is
$\mathbf{S}_e=\mathbf{S}_{e,J_e}$.

For unit-norm keys, scalar erasure satisfies
$\|\mathbf{I}-\beta\mathbf{k}\mathbf{k}^{\top}\|_2\le 1$.
Together with the contractive entry-level decay, this makes the direct
propagation of the previous state non-expansive for fixed keys and gates.
Further update details are in Appendix~\ref{app:gated_delta}.

\subsection{Memory-Augmented Action Generation}
At each denoising iteration, the action DiT uses its features
to retrieve context from short- and long-term memories,
as shown in Figure~\ref{fig:architecture}(b).
For visual memory, queries from action features perform
comparably to those from current-observation features in our
experiments. We choose the former because it avoids recursively encoding
historical observations with memory and reduces training cost.

\noindent\textbf{Short-term memory retrieval.}
At layer $r$, action queries attend to the KV representations
of the current observation, the permanent visual anchor, the visual
FIFO, and the noisy action sequence through the backbone's MoT attention.
A shared cross-attention module reads the interaction
content tokens $\mathbf{Z}_i$ every five
layers of the action DiT. Relative-age embeddings augment historical visual keys and
interaction content tokens at retrieval time, allowing the readers to
distinguish entries within each FIFO.

\noindent\textbf{Long-term memory retrieval.}
Let $\mathbf{S}_{\mathrm{v}}^{(r)}$ and $\mathbf{S}_{\mathrm{a}}$ denote
the visual and interaction LTM states in the memory bank $\mathcal{M}_t$,
with head indices omitted. For one head of either memory, the readout is
\begin{equation}
\begin{aligned}
    \mathbf{q}
        &= \frac{\mathbf{W}_q\mathbf{x}}
           {\sqrt{d_k}\,\|\mathbf{W}_q\mathbf{x}\|_2}, \\
    \mathbf{u}
        &= \mathbf{S}^{\top}\mathbf{q}.
\end{aligned}
\end{equation}
Here, $\mathbf{x}$ is the action feature supplied to the reader,
$\mathbf{W}_q$ is its learned query projection, and $\mathbf{S}$ is
the corresponding per-head state.
Thus, $\mathbf{q}\in\mathbb{R}^{d_k}$ and
$\mathbf{u}\in\mathbb{R}^{d_v}$. The head outputs are concatenated,
layer-normalized, and projected back to the hidden dimension of the action DiT.

The visual LTM reader uses the attention block's normalized,
timestep-modulated input. Its projected readout is added to the MoT
attention output before the timestep-conditioned residual gate.
The interaction LTM reader uses normalized action features after
interaction STM cross-attention and adds its output as a separate
residual every five layers. Visual readers have
layer-specific projections, while the interaction reader is shared
across insertion layers.

\noindent\textbf{Training.}
We follow Fast-WAM's joint flow-matching formulation
\cite{lipman2023flow,yuan2026fastwam}.
Let $\mathbf{y}$ denote either a normalized demonstrated action
chunk or future video latents. For each branch, we independently
sample a noise level $\tau\in(0,1)$ and standard Gaussian noise
$\boldsymbol{\epsilon}$ to construct
\begin{equation}
    \mathbf{y}_{\tau}
        = (1-\tau)\mathbf{y}+\tau\boldsymbol{\epsilon}.
\end{equation}
The corresponding branch predicts the velocity
$\boldsymbol{\epsilon}-\mathbf{y}$, with objective
\begin{equation}
    \mathcal{L}_{\mathrm{FM}}(\mathbf{y})
    = \mathbb{E}\!\left[
      \left\|f_{\theta}(\mathbf{y}_{\tau},\tau, o_t,\ell,\mathcal{M}_t)
      -(\boldsymbol{\epsilon}-\mathbf{y})\right\|_2^2
      \right].
\end{equation}
Here, $f_{\theta}$ denotes the corresponding branch's velocity predictor.
The current observation $o_t$ provides clean conditioning alongside
the instruction $\ell$. In implementation, we apply noise-level weighting
and average squared errors over non-padded elements.
Applying this objective to actions and video latents gives
$\mathcal{L}_{\mathrm{act}}$ and $\mathcal{L}_{\mathrm{vid}}$,
respectively. The combined objective is
\begin{equation}
    \mathcal{L}_{\mathrm{train}}
        = \mathcal{L}_{\mathrm{act}}
          +\lambda_{\mathrm{vid}}\mathcal{L}_{\mathrm{vid}}.
\end{equation}
Here, $\lambda_{\mathrm{vid}}$ balances action learning and video co-training.

During training, STM contains the preceding history window,
while LTM is reconstructed by replaying earlier sampled history
from the same episode under current parameters.
We detach historical visual KV while keeping memory
writers, readers, and the executed-prefix encoder trainable.
Backpropagating through historical visual encoding increases training
time and GPU memory substantially, with little performance
change in our experiments.

Additionally, to discourage shortcuts from visual context and action
continuity, we randomly mask visual STM entries at retrieval and
add Gaussian noise to selected dimensions of normalized executed
prefixes before encoding.

\noindent\textbf{Closed-loop inference.}
During inference, the memory bank $\mathcal{M}_t$ remains fixed
throughout action denoising. After generation, sampled observations
enter the visual FIFO, while action prefixes enter the interaction
FIFO once executed, with evicted entries consolidated into LTM.
Implementation and training details are in Appendix~\ref{app:implementation}.

%% file: sec/4_experiments.tex
\section{Experiments}
\label{sec:experiments}

\begin{figure*}[t]
    \centering
    \includegraphics[width=\textwidth]{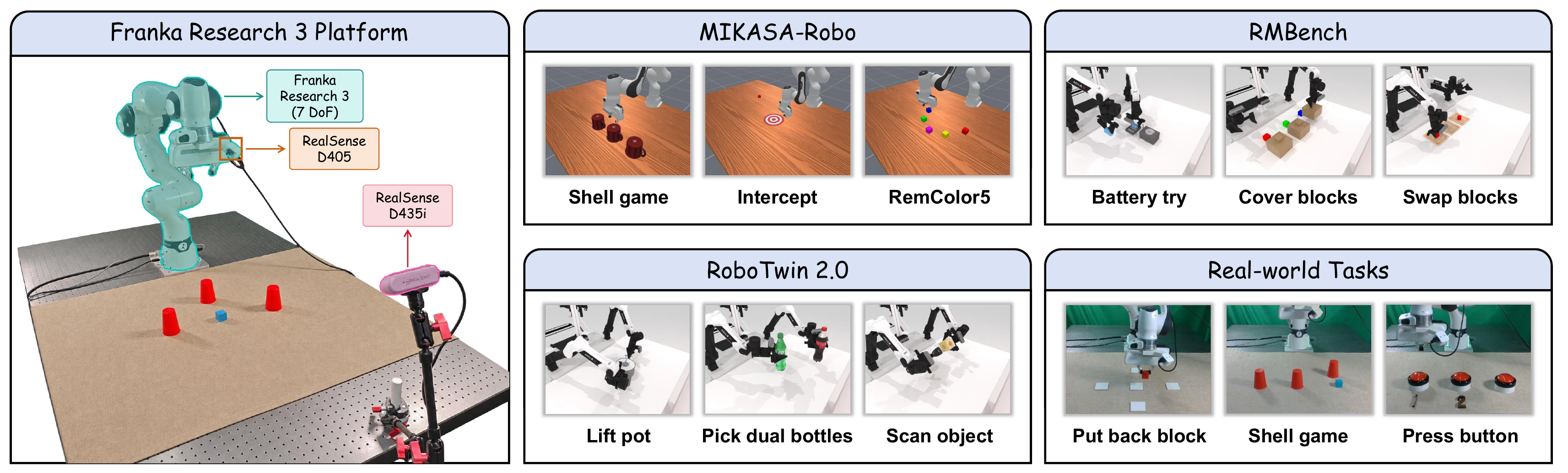}
    \caption{Robot platform used in real-world experiments and representative
    tasks from simulation benchmarks and real-world evaluation.}
    \label{fig:experimental_setup}
\end{figure*}

\input{tables/results_mikasa}
\input{tables/results_rmbench}

\subsection{Experimental Setup}

We evaluate M\textsuperscript{2}-WAM on three simulation benchmarks and
three real-world tasks to assess its use of historical context and
general manipulation performance. Figure~\ref{fig:experimental_setup}
shows the robot platform and representative tasks.

\noindent\textbf{Simulation benchmarks.}
\textbf{MIKASA-Robo} \cite{cherepanov2026memory} tests abilities such as remembering
occluded object locations, recalling previously observed colors, and
inferring object motion from observation sequences. We select
ShellGameTouch, InterceptMedium, and RememberColor3/5/9 to assess these
abilities over relatively short task horizons.
\textbf{RMBench} \cite{chen2026rmbench} comprises nine memory-dependent tasks requiring
recall of earlier observations or accumulation of past interaction
outcomes. We use it to evaluate whether memory supports longer,
multi-stage manipulation, including repeated attempts and tracking
task progress.
\textbf{RoboTwin~2.0} \cite{chen2025robotwin} provides diverse bimanual manipulation
tasks with variations in scene appearance and clutter. Its Clean and
Randomized settings allow us to assess whether memory augmentation
preserves the backbone's performance on general manipulation tasks
with less dependence on historical context.

\noindent\textbf{Real-world setup.}
We further evaluate on three memory-dependent real-world tasks:
Put Back Block, Shell Game, and Press Button.
We use a \mbox{7-DoF} Franka Research~3 arm with a main-view
RealSense D435i camera and a wrist-mounted RealSense D405 camera. The action space consists of end-effector pose increments
and a binary gripper open/close command.
We run inference on one NVIDIA RTX~3090 GPU and send
control commands at 15~Hz.
For each task, we collected 50 demonstration trajectories
using an Omega.7 teleoperation device and trained a separate policy
for up to 20,000 optimization steps.

\noindent\textbf{Evaluation protocol.}
For our main simulation experiments, we follow the official evaluation
protocols with 100 episodes per task. We evaluate each real-world policy
over 50 consecutive trials per task. We report task success rates and their
unweighted average across tasks, with the two RoboTwin~2.0 settings
reported separately.

\subsection{Main Results}
\label{sec:main_results}

We first report simulation results for memory-dependent and general
manipulation tasks. We build M\textsuperscript{2}-WAM on Fast-WAM
\cite{yuan2026fastwam} and compare the two models to evaluate the benefits
of our memory mechanism.

\noindent\textbf{MIKASA-Robo.}
M\textsuperscript{2}-WAM improves performance on short-horizon tasks
requiring visual recall and motion inference.
In Table~\ref{tab:mikasa}, it achieves the highest average success
among the compared methods at 78.8\%, exceeding Fast-WAM
and SimpleMemVLA by 53.2 and 4.8 percentage points, respectively.
Relative to SimpleMemVLA, the largest gains occur on RememberColor3
and InterceptMedium, while improvements also extend to RememberColor5/9.
This pattern suggests benefits for both recalling earlier visual
attributes and using observation sequences to infer motion.

\noindent\textbf{RMBench.}
Memory augmentation also yields substantial gains on longer,
multi-stage manipulation tasks.
M\textsuperscript{2}-WAM achieves 65.78\% average success, the highest
among the methods compared in Table~\ref{tab:rmbench}. This exceeds DRAM
by 8.00 percentage points and Fast-WAM by 61.22 percentage points.
The largest gains over DRAM occur on SwapT, PutBack and BlocksRank. The improvement thus
spans both recall of an earlier scene and decisions requiring
accumulated history.
Success remains low on ObsPick and PressButton. Our failure-case
analysis suggests that limitations in the Fast-WAM backbone's generalization
and character recognition may contribute to these errors.

\input{tables/results_robotwin}

\noindent\textbf{RoboTwin~2.0.}
The results in Table~\ref{tab:robotwin} show that memory augmentation
largely preserves Fast-WAM's performance on general manipulation tasks.
M\textsuperscript{2}-WAM achieves 91.78\%
and 90.70\% success in the Clean and Randomized settings, respectively.
Together with the gains on memory-dependent benchmarks, these results
suggest that memory augmentation improves history-dependent decisions
without substantially compromising general manipulation performance.

\begin{figure}[!b]
    \centering
    \includegraphics[width=0.85\linewidth]{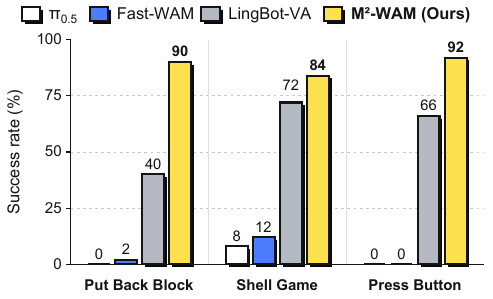}
    \caption{\textbf{Results on real-world tasks.}
    M\textsuperscript{2}-WAM achieves the highest success on all three tasks,
    with the largest margin over LingBot-VA on Put Back Block.}
    \label{fig:real_world_results}
\end{figure}

\noindent\textbf{Real-world experiments.}
Figure~\ref{fig:real_world_results} shows that the gains on
memory-dependent tasks extend to the physical robot.
M\textsuperscript{2}-WAM achieves
88.7\% average success, exceeding LingBot-VA, the strongest baseline,
by 29.3 percentage points. The improvement holds across all three
tasks.
Relative to Fast-WAM, the average gain is 84.0 percentage points,
supporting the benefit of augmenting the backbone with historical
context for these tasks.

\begin{figure}[t]
    \centering
    \includegraphics[width=\linewidth]{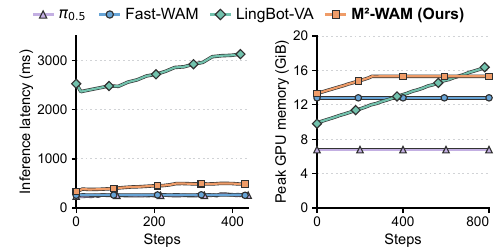}
    \caption{\textbf{Inference efficiency as trajectories grow.}
    Inference latency (left) and peak GPU memory usage (right).
    Both costs stabilize for M\textsuperscript{2}-WAM as history accumulates.}
    \label{fig:policy_efficiency}
\end{figure}

\subsection{Memory Design Ablations}
\label{sec:memory_ablations}

We compare five memory variants on MIKASA-Robo and RMBench
in Table~\ref{tab:mikasa} and Table~\ref{tab:rmbench} to assess memory
timescales, modality-specific processing, and history consolidation.
All variants share the full model's training data and optimization budget.

\noindent\textbf{Memory hierarchy.}
The full hierarchy outperforms either memory timescale alone
on average across benchmarks.
\emph{w/o-LTM} retains the STMs and visual anchor, whereas
\emph{w/o-STM} stores history only in associative states.
Their RMBench averages fall from 65.78\% to 38.89\% and 22.33\%,
respectively, supporting the combination of explicit recent context
and compressed longer-term history.
Without LTM, success on the four $M(n)$ tasks drops from 49.0\%
to 9.75\%, highlighting the importance of retention beyond the
recent window.
On MIKASA-Robo, removing LTM lowers the
RememberColor3/5/9 average from 68.3\% to 45.0\%, although
ShellGameTouch remains competitive.

\noindent\textbf{Modality-specific memory.}
Interaction history complements visual memory, and modality separation
also matters.
Removing interaction memory reduces RMBench success to 14.33\%.
\emph{w/o modality separation} retains both visual and interaction
histories but projects executed-prefix tokens into visual KV space
for shared storage and retrieval. Success recovers to 32.22\%,
but remains well below the full model.
Both variants also underperform on MIKASA-Robo, with smaller losses.
These comparisons favor retaining behavioral history through a
dedicated interaction-memory pathway alongside visual memory.

\noindent\textbf{Uncompressed history.}
Retaining more history alone does not ensure better decisions.
\emph{Full-history memory} stores all sampled visual KV entries
and executed prefixes without eviction or associative consolidation.
It achieves 59.8\% on MIKASA-Robo and only 3.89\% on RMBench,
well below the full model on both benchmarks.
The RMBench result is also slightly below Fast-WAM (4.56\%).
Accumulated historical tokens may dilute attention to current
observations and make task-relevant
context harder to retrieve.

\subsection{In-depth Analysis}

\subsubsection{Inference Efficiency and Scalability}
\label{sec:inference_efficiency}

Figure~\ref{fig:policy_efficiency} shows that M\textsuperscript{2}-WAM
adds bounded inference overhead to Fast-WAM.
After an initial increase, latency and GPU memory usage stabilize
as trajectories grow: bounded FIFOs and fixed-size associative states
incorporate new history without expanding storage or retrieval context.
For real-world inference, both models use precomputed text embeddings
without loading the text encoder. For GPU memory measurements, we
replace LingBot-VA's default KV cache preallocation with on-demand
allocation.

\subsubsection{Effect of Memory Retrieval on Decision-Making}
\label{sec:memory_retrieval_effect}

\begin{figure}[!t]
    \centering
    \includegraphics[width=0.95\linewidth]{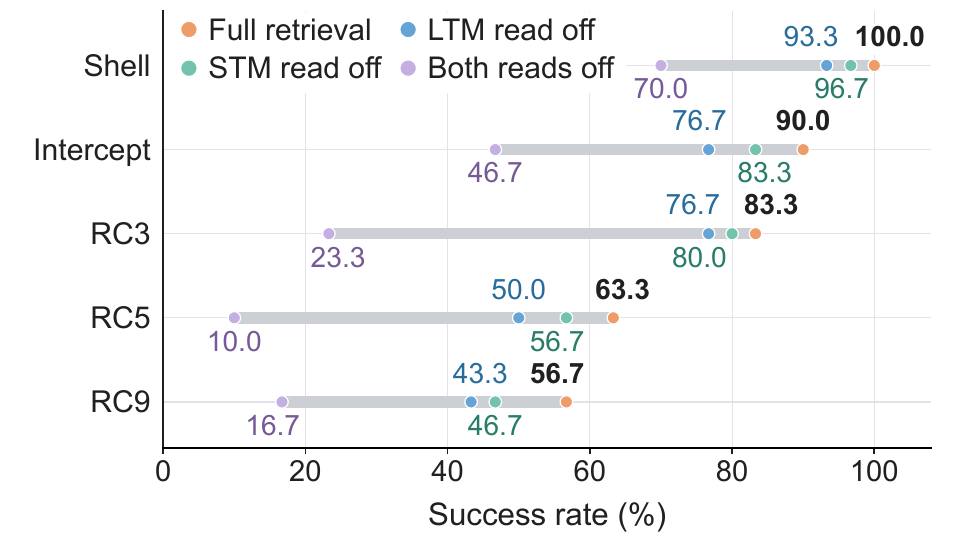}
    \caption{\textbf{Effect of memory retrieval on task success.}
    Full retrieval performs best on all five MIKASA-Robo tasks,
    while disabling both memory levels causes the largest decline.}
    \label{fig:memory_retrieval_intervention}
\end{figure}

For each task, we compare full retrieval with STM retrieval,
LTM retrieval, or both disabled, using a fixed checkpoint and
30 matched-seed rollouts per condition.
All interventions affect visual and interaction memory while
retaining the initial visual anchor.
In Figure~\ref{fig:memory_retrieval_intervention}, full retrieval
performs best on all five MIKASA-Robo tasks.
Disabling both particularly degrades RememberColor performance,
while retaining either level recovers much of the lost performance.
Disabling LTM consistently causes larger drops than disabling STM,
highlighting the contribution of consolidated history.
Across the RememberColor tasks, combining both memory levels
provides increasing gains over LTM retrieval alone as the number
of candidate colors grows, suggesting complementary contributions.

\subsubsection{Qualitative Analysis}
\label{sec:qualitative_analysis}

\begin{figure}[!htbp]
    \centering
    \includegraphics[width=\linewidth]{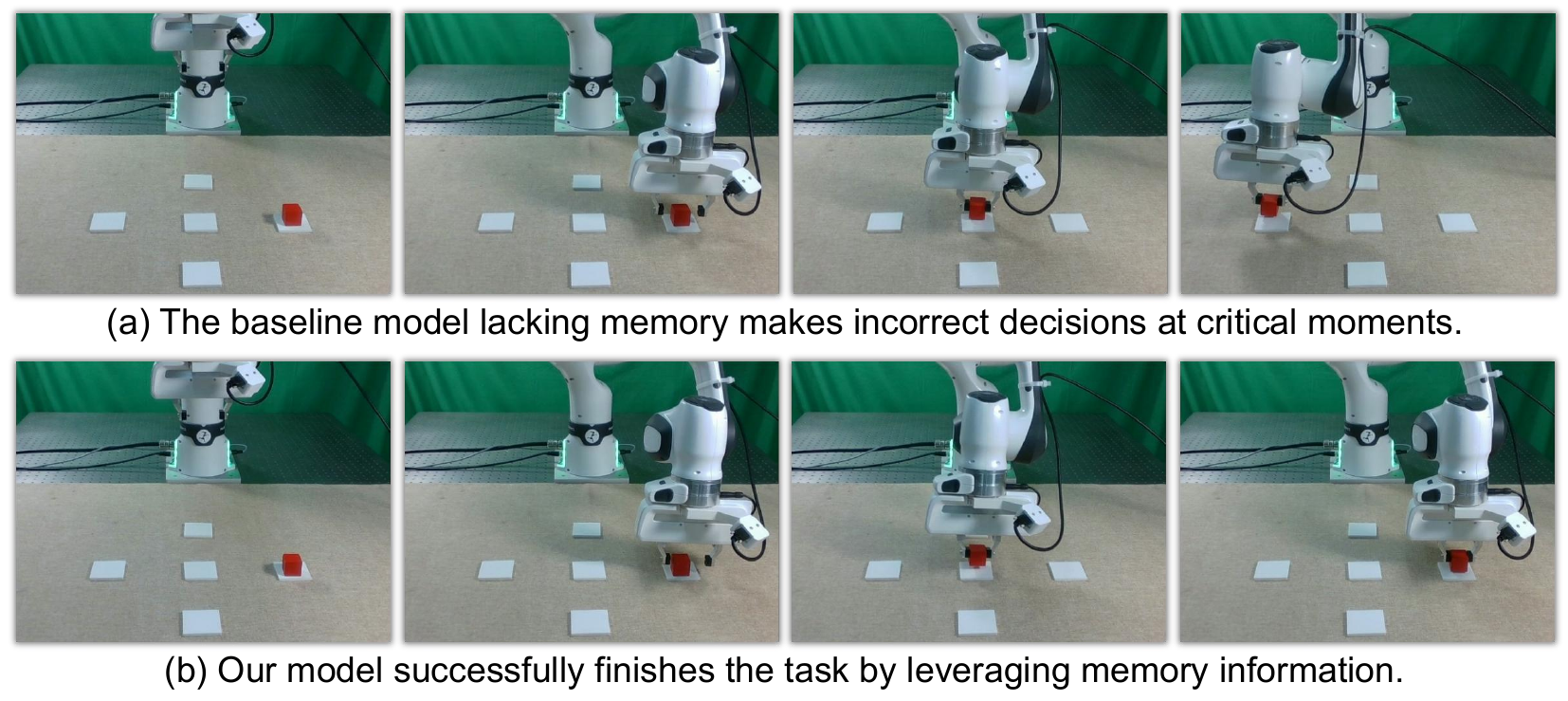}
    \caption{\textbf{Qualitative comparison on Put Back Block.}
    (a) Fast-WAM selects an incorrect placement location.
    (b) M\textsuperscript{2}-WAM selects the correct target and succeeds.}
    \label{fig:real_world_case}
\end{figure}

Figure~\ref{fig:real_world_case} compares the policies on real-world
Put Back Block. Both grasp the block, but Fast-WAM selects an incorrect
placement location, whereas M\textsuperscript{2}-WAM reaches the correct target.
This contrast suggests that historical context supports target selection
after grasping.
Additional qualitative examples are in
Appendix~\ref{app:qualitative}.

%% file: tables/results_mikasa.tex
\begin{table*}[t]
\centering
\caption{\textbf{MIKASA-Robo success rates (\%).}
RC3/5/9 denote RememberColor with three, five, or nine candidates.
Bold marks column-wise best results in the main comparison.
Parentheses show gains over Fast-WAM in percentage points.}
\label{tab:mikasa}
\footnotesize
\setlength{\tabcolsep}{4pt}
\renewcommand{\arraystretch}{1.05}
\begin{tabular*}{0.8\textwidth}{@{\extracolsep{\fill}}lcccccc@{}}
\toprule
Method & ShellGameTouch & InterceptMedium & RC3 & RC5 & RC9 & Avg. $\uparrow$ \\
\midrule
$\pi_0$ \cite{black2024pi_0} & 33 & 42 & 35 & 22 & 15 & 29.4 \\
MemoryVLA \cite{shi2025memoryvla} & 88 & 24 & 44 & 30 & 20 & 41.2 \\
MemoryVLA++ \cite{shi2026memoryvla++} & 97 & 40 & 50 & 19 & 16 & 44.4 \\
GMP \cite{gao2026gatedmemorypolicy} & 98 & 83 & 80 & 61 & 17 & 67.8 \\
SimpleMemVLA \cite{yin2026simplememvla} & \textbf{99} & 83 & 71 & 58 & 59 & 74.0 \\
\fastwamrow{}
Fast-WAM \cite{yuan2026fastwam} & 30 & 38 & 28 & 20 & 12 & 25.6 \\
\midrule
\oursrow{}
\textbf{M\textsuperscript{2}-WAM (Ours)} & 98 & \textbf{91} & \textbf{81} & \textbf{63} & \textbf{61} & \textbf{78.8}\avgdelta{+53.2} \\
\midrule
\resultrow{black!8}{}
\multicolumn{7}{l}{\textit{Memory variants}} \\
w/o-LTM & 99 & 83 & 47 & 43 & 45 & 63.4 \\
w/o-STM & 96 & 68 & 64 & 47 & 49 & 64.8 \\
Full-history memory & 91 & 69 & 49 & 40 & 50 & 59.8 \\
w/o modality separation & 94 & 81 & 55 & 61 & 58 & 69.8 \\
w/o-interaction memory & 84 & 63 & 68 & 58 & 59 & 66.4 \\
\bottomrule
\end{tabular*}
\end{table*}

%% file: tables/results_rmbench.tex
\begin{table*}[t]
\centering
\caption{\textbf{RMBench success rates (\%).}
$M(1)$ and $M(n)$ require one and multiple task-relevant past observations,
respectively~\cite{chen2026rmbench}.
Boldface and parentheses follow Table~\ref{tab:mikasa}.}
\label{tab:rmbench}
\footnotesize
\setlength{\tabcolsep}{1.8pt}
\renewcommand{\arraystretch}{1.05}
\begin{tabular*}{\textwidth}{@{\extracolsep{\fill}}lcccccccccc@{}}
\toprule
\multicolumn{1}{c}{\multirow{2}{*}{Method}} & \multicolumn{5}{c}{$M(1)$: one past observation} & \multicolumn{4}{c}{$M(n)$: accumulated history} & \multirow{2}{*}{Avg. $\uparrow$} \\
\cmidrule(lr){2-6}\cmidrule(lr){7-10}
& ObsPick & Rearrange & PutBack & SwapBlocks & SwapT & BatteryTry & BlocksRank & CoverBlocks & PressButton & \\
\midrule
$\pi_{0.5}$ \cite{intelligence2025pi_} & 9 & 13 & 11 & 24 & 15 & 16 & 6 & 0 & 0 & 10.44 \\
X-VLA \cite{zheng2025x} & 9 & 13 & 18 & 16 & 3 & 26 & 1 & 2 & 0 & 9.78 \\
Mem-0 \cite{chen2026rmbench} & 4 & 89 & 90 & 67 & 14 & 28 & 18 & 68 & 0 & 42.00 \\
HiMem-WAM \cite{sun2026himemwam} & \textbf{28} & 33 & 32 & 38 & 27 & 28 & 24 & 19 & 8 & 26.33 \\
DRAM \cite{zhao2026dram} & 15 & 95 & 65 & \textbf{100} & 45 & \textbf{65} & 45 & 70 & \textbf{20} & 57.78 \\
\fastwamrow{\textwidth}
Fast-WAM \cite{yuan2026fastwam} & 1 & 20 & 0 & 4 & 11 & 2 & 3 & 0 & 0 & 4.56 \\
\midrule
\oursrow{\textwidth}
\textbf{M\textsuperscript{2}-WAM (Ours)} & 11 & \textbf{97} & \textbf{100} & \textbf{100} & \textbf{88} & 28 & \textbf{76} & \textbf{74} & 18 & \textbf{65.78}\avgdelta{+61.22} \\
\midrule
\resultrow{black!8}{\textwidth}
\multicolumn{11}{l}{\textit{Memory variants}} \\
w/o-LTM & 8 & 87 & 100 & 53 & 63 & 20 & 19 & 0 & 0 & 38.89 \\
w/o-STM & 3 & 56 & 32 & 68 & 19 & 17 & 6 & 0 & 0 & 22.33 \\
Full-history memory & 10 & 0 & 0 & 0 & 20 & 5 & 0 & 0 & 0 & 3.89 \\
w/o modality separation & 10 & 10 & 56 & 98 & 64 & 16 & 8 & 24 & 4 & 32.22 \\
w/o-interaction memory & 8 & 19 & 0 & 19 & 38 & 12 & 33 & 0 & 0 & 14.33 \\
\bottomrule
\end{tabular*}
\end{table*}

%% file: tables/results_robotwin.tex
\begin{table}[!ht]
\centering
\caption{\textbf{RoboTwin~2.0 success rates (\%).}
Avg. is the mean of the two settings.}
\label{tab:robotwin}
\footnotesize
\setlength{\tabcolsep}{3pt}
\renewcommand{\arraystretch}{1.05}
\begin{tabular*}{\columnwidth}{@{\extracolsep{\fill}}lccc@{}}
\toprule
Method & Clean & Randomized & Avg. $\uparrow$ \\
\midrule
$\pi_0$ \cite{black2024pi_0} & 65.92 & 58.40 & 62.16 \\
X-VLA \cite{zheng2025x} & 72.88 & 72.84 & 72.86 \\
$\pi_{0.5}$ \cite{intelligence2025pi_} & 82.74 & 76.76 & 79.75 \\
Motus \cite{bi2025motusunifiedlatentaction} & 88.66 & 87.02 & 87.84 \\
LingBot-VA \cite{lingbot-va2026} & \textbf{92.90} & \textbf{91.50} & \textbf{92.20} \\
\fastwamrow{\columnwidth}
Fast-WAM \cite{yuan2026fastwam} & 91.62 & 91.24 & 91.43 \\
\midrule
\oursrow{\columnwidth}
\textbf{M\textsuperscript{2}-WAM (Ours)} & 91.78 & 90.70 & 91.24 \\
\bottomrule
\end{tabular*}
\end{table}

%% file: sec/5_conclusion.tex
\section{Conclusion}
\label{sec:conclusion}

We present M\textsuperscript{2}-WAM, a world-action model with
hierarchical visual and interaction memories.
The two pathways preserve dense visual context and encode executed
action prefixes into compact temporal representations, respectively.
Recent entries are retained explicitly, while older ones are consolidated
into fixed-size associative states through adapted gated delta updates.
Layer-wise retrieval brings both memory levels into action generation
with bounded online storage and retrieval costs.
Simulation and real-world experiments show improved memory-dependent
manipulation while largely preserving general manipulation performance.
These findings support modality-specific memory organization and
retrieval across timescales as a promising direction for memory-augmented
world-action modeling.

%% file: sec/X_suppl.tex
\clearpage
\setcounter{page}{1}
\renewcommand{\thepage}{S\arabic{page}}
\maketitlesupplementary
\appendix

\section*{Contents}
\begingroup
\setcounter{tocdepth}{2}
\makeatletter
\@starttoc{stoc}
\makeatother
\endgroup
\let\supplementOriginalAddcontentsline\addcontentsline
\renewcommand{\addcontentsline}[3]{%
    \ifstrequal{#1}{toc}{%
        \supplementOriginalAddcontentsline{stoc}{#2}{#3}%
    }{%
        \supplementOriginalAddcontentsline{#1}{#2}{#3}%
    }%
}

\section{Implementation and Training Details}
\label{app:implementation}

This section describes the architecture and hyperparameters of
M\textsuperscript{2}-WAM and explains how memory is constructed
during training and online inference. We then detail the gated delta
updates and analyze the storage and retrieval costs of the
memory hierarchy.

\subsection{Architecture and Hyperparameters}
\label{app:architecture}

We retain the video and action DiT architectures of Fast-WAM
\cite{yuan2026fastwam} and add the memory modules described in
Section~\ref{sec:method}. Table~\ref{tab:appendix_shared_hyperparameters}
summarizes the shared architecture, optimization, and inference
settings; Table~\ref{tab:appendix_task_hyperparameters} lists
the task-specific settings. The permanent visual anchor is
stored separately and does not occupy any of the $N$ FIFO slots.

We train our policies on PPU-ZW810E accelerators, with training
resources summarized in Table~\ref{tab:appendix_task_hyperparameters}.
Simulation inference uses NVIDIA RTX PRO 5000 Blackwell GPUs,
while real-world deployment uses a single NVIDIA RTX 3090.

\begin{table*}[t]
    \centering
    \caption{\textbf{Shared architecture, training, and inference hyperparameters.}
    Noise parameters apply only when
    executed-prefix augmentation is enabled.}
    \label{tab:appendix_shared_hyperparameters}
    \small
    \begin{tabular}{ll}
        \toprule
        Parameter & Value \\
        \midrule
        \multicolumn{2}{l}{\textbf{Base architecture}} \\
        Layers (video / action) & 30 / 30 \\
        Hidden dimension (video / action) & 3,072 / 1,024 \\
        FFN dimension (video / action) & 14,336 / 4,096 \\
        Attention heads / head dimension & 24 / 128 in both branches \\
        Proprioceptive projection & Linear, output dimension 4,096 \\
        Text conditioning & Precomputed embeddings \\
        VAE parameters & Frozen \\
        \midrule
        \multicolumn{2}{l}{\textbf{Memory modules}} \\
        LTM heads (visual / interaction) & 24 per layer / 4 in one shared state \\
        Per-head key/value dimensions $(d_k,d_v)$ & $(128,128)$ \\
        Writer feature dimension $d_w$ & 256 \\
        Prefix encoder convolutions & Three Conv1d layers, 256 channels each \\
        Kernel sizes / padding & $(1,3,3)$ / $(0,1,1)$ \\
        Encoder activation & SiLU after each convolution \\
        Encoder output & Adaptive average pooling, Linear, LayerNorm \\
        Prefix tokens $P$ / dimension $d_m$ & 4 / 1,024 \\
        Interaction reader placement & Every 5 action DiT layers; shared parameters \\
        Reader output projection initialization & Zero for LTM and interaction STM readers \\
        \midrule
        \multicolumn{2}{l}{\textbf{Optimization}} \\
        Optimizer / moment coefficients & AdamW / $(0.9,0.95)$ \\
        Peak learning rate / weight decay & $10^{-4}$ / $0.01$ \\
        Learning-rate schedule & 5\% linear warmup, then cosine decay \\
        Final learning rate & 1\% of peak learning rate \\
        Gradient clipping & Global norm 1.0 \\
        Video loss weight $\lambda_{\mathrm{vid}}$ & 1 \\
        Noise-schedule shift (video / action) & 5 / 5 \\
        \midrule
        \multicolumn{2}{l}{\textbf{Memory augmentation (training only)}} \\
        Visual STM dropout probability & 0.25, per observation entry \\
        Executed-prefix noise standard deviation & 0.005, in normalized action space \\
        Executed-prefix noise probability & 0.5 per training sample \\
        \midrule
        \multicolumn{2}{l}{\textbf{Inference}} \\
        Action denoising steps & 10 \\
        \bottomrule
    \end{tabular}

\end{table*}

\begin{algorithm}[t]
    \caption{Online memory maintenance}
    \label{alg:online_memory}
    \small
    \begin{algorithmic}[1]
        \Require Policy $\pi_\theta$, instruction $\ell$, settings in
        Table~\ref{tab:appendix_task_hyperparameters}
        \State $\mathcal{Q}_{\mathrm{v}},\mathcal{Q}_{\mathrm{a}}\gets [\,],[\,]$;
        $\mathcal{C}_{\mathrm{anchor}}\gets\varnothing$
        \State $\mathbf{S}_{\mathrm{v}}^{(r)}\gets\mathbf{0}$ for all $r$;
        $\mathbf{S}_{\mathrm{a}}\gets\mathbf{0}$
        \For{each replanning step $t$ until episode termination}
            \State Observe $o_t$ and compute $\mathcal{C}_t=\{(\mathbf{K}_t^{(r)},\mathbf{V}_t^{(r)})\}_r$
            \State $\mathcal{M}_t\gets(\mathcal{C}_{\mathrm{anchor}},\mathcal{Q}_{\mathrm{v}},\mathcal{Q}_{\mathrm{a}},
                \{\mathbf{S}_{\mathrm{v}}^{(r)}\}_r,\mathbf{S}_{\mathrm{a}})$
            \State $\mathbf{A}_t\sim\pi_\theta(\cdot\mid o_t,\ell,\mathcal{M}_t)$ \Comment{Fixed $\mathcal{M}_t$}
            \If{$o_t$ is selected by the memory sampling stride}
                \If{$|\mathcal{Q}_{\mathrm{v}}|=N$}
                    \State \Call{Consolidate}{$\mathcal{Q}_{\mathrm{v}}.\operatorname{front}(),\mathcal{Q}_{\mathrm{a}}.\operatorname{front}()$}
                \EndIf
                \State $\mathcal{Q}_{\mathrm{v}}.\operatorname{push}(\mathcal{C}_t)$
                \State $\mathcal{C}_{\mathrm{anchor}}\gets\mathcal{C}_t$ if $\mathcal{C}_{\mathrm{anchor}}=\varnothing$
            \EndIf
            \State Execute prefix $\mathbf{A}_t^{\mathrm{exec}}$ of $\mathbf{A}_t$
            \If{$o_t$ was selected and the prefix is complete}
                \State $\mathcal{Q}_{\mathrm{a}}.\operatorname{push}(\mathbf{A}_t^{\mathrm{exec}})$
            \EndIf
        \EndFor
    \end{algorithmic}
\end{algorithm}

\subsection{Training and Online Memory Construction}
\label{app:memory_construction}

\noindent\textbf{Training memory construction.}
For each training sample at replanning step $t$, we construct memory
from earlier observations and demonstrated action prefixes in the
same episode. Each sampled observation is paired with the subsequent
action prefix, all of whose commands precede the current prediction.
The most recent $N$ sampled entries populate the visual and
interaction FIFOs, with fewer entries available near the episode start.
Earlier entries are replayed chronologically from zero-initialized
LTM states under the current model parameters. The first observation
also provides the permanent visual anchor. Memory is reconstructed
for each sample from its preceding history, rather than carried
across optimization steps.

\noindent\textbf{Gradient handling.}
Historical observations are encoded independently, and their visual
KV representations are detached before memory construction.
Gradients propagate through the memory writers, readers,
executed-prefix encoder, and replayed LTM updates.
Both DiTs are trained through the current sample's objectives,
while historical visual encoding is excluded from the backward pass.

\noindent\textbf{Memory augmentation.}
Visual STM dropout masks complete observation entries at retrieval
while retaining the permanent anchor. A common mask is used across
layers and heads. Stored entries and LTM writer inputs are unchanged.
When executed-prefix augmentation is enabled, each training sample
is selected with the probability in
Table~\ref{tab:appendix_shared_hyperparameters}, and zero-mean
Gaussian noise is added to selected dimensions of its normalized
executed prefixes before encoding.
For joint-space actions in RMBench and RoboTwin~2.0, we perturb
arm joint dimensions while leaving gripper commands and padded
timesteps unchanged. Both augmentations are disabled during evaluation.

\begin{table*}[t]
    \centering
    \caption{\textbf{Task-specific hyperparameters.} The sampling stride counts
    source frames; prediction and execution lengths count action steps.
    FIFO capacity excludes the permanent visual anchor.}
    \label{tab:appendix_task_hyperparameters}
    \small
    \begin{tabular}{lcccc}
        \toprule
        Parameter & MIKASA-Robo & RMBench & RoboTwin 2.0 & Real world \\
        \midrule
        FIFO capacity $N$ & 4 & 16 & 16 & 16 \\
        Memory sampling stride (source frames) & 4 & 16 & 16 & 16 \\
        Prediction horizon $H$ (action steps) & 8 & 32 & 32 & 32 \\
        Executed prefix length (action steps) & 4 & 16 & 16 & 16 \\
        Batch size per device & 4 & 2 & 4 & 4 \\
        Training PPUs & 32 & 64 & 64 & 16 \\
        Global batch size & 128 & 128 & 256 & 64 \\
        Approximate training cost (PPU-hours) & 1,660 & 1,890 & 2,760 & --- \\
        \bottomrule
    \end{tabular}
\end{table*}

\noindent\textbf{Online memory construction.}
Algorithm~\ref{alg:online_memory} summarizes online memory maintenance
for the full model.
The memory bank $\mathcal{M}_t$ remains fixed throughout action
denoising, and updates become available at the next replanning step.
Sampled visual entries are committed after action generation,
and their interaction entries after the corresponding prefix has
been executed. The two FIFOs are aligned by source observation
before the next prediction. Only complete executed prefixes are
stored; unexecuted action tails and incomplete terminal prefixes
are omitted.

In the algorithm, $\mathcal{Q}_{\mathrm{v}}$ and $\mathcal{Q}_{\mathrm{a}}$
denote the visual and interaction FIFOs. The operation
$\operatorname{push}$ appends an entry and evicts the oldest if
capacity $N$ is exceeded; $\operatorname{front}$ returns the oldest
entry without removing it.
$\mathcal{C}_t$ denotes the current layer-wise visual KV, and
$\mathcal{C}_{\mathrm{anchor}}$ stores the permanent anchor.
\textsc{Consolidate} updates the visual and interaction LTM states
in place from the oldest paired entries, following
Appendix~\ref{app:gated_delta}. Each entry retains its source-frame
metadata, omitted from the pseudocode for readability.
Action prefixes are normalized before storage.

\subsection{Gated Delta Update Details}
\label{app:gated_delta}

This subsection expands the gated delta update in
Section~\ref{sec:associative_ltm}.
As in the main text, $e$ indexes evicted entries, each containing
$J_e$ tokens indexed by $j=1,\ldots,J_e$.
Interaction entries contain $J_e=P$ tokens.
We describe one head with state
$\mathbf{S}\in\mathbb{R}^{d_k\times d_v}$, omitting modality,
layer, and head indices.

\noindent\textbf{Gate parameterization.}
Let $\mathbf{h}_{e,j}\in\mathbb{R}^{d_w}$ denote the writer feature,
where $d_w$ is the writer dimension in
Table~\ref{tab:appendix_shared_hyperparameters}.
Visual writer features are obtained by normalizing and projecting
the video DiT's values. For an interaction entry $e$ paired with
observation $i$, the writer normalizes and projects $\mathbf{Z}_i$,
then adds the sinusoidal within-chunk phase encoding before
predicting keys, values, and gates.
FIFO-relative age embeddings are applied only at retrieval.
Separate bias-free projections predict the erase and write gates:
\begin{equation}
\begin{aligned}
    \beta_{e,j} &= \operatorname{sigmoid}
        (\mathbf{p}_{\beta}^{\top}\mathbf{h}_{e,j}), \\
    \mathbf{w}_{e,j} &= \operatorname{sigmoid}
        (\mathbf{W}_{w}\mathbf{h}_{e,j}).
\end{aligned}
\end{equation}
Here, $\mathbf{p}_{\beta}\in\mathbb{R}^{d_w}$ and
$\mathbf{W}_{w}\in\mathbb{R}^{d_v\times d_w}$ are learned parameters
for the selected head, producing one erase scalar and $d_v$ write
coefficients per token. Decay rates instead use an entry-averaged feature:
\begin{equation}
\begin{aligned}
    \bar{\mathbf{h}}_e &= \frac{1}{J_e}
        \sum_{j=1}^{J_e}\mathbf{h}_{e,j}, \\
    \boldsymbol{\lambda}_e &= \exp(a)\,
        \operatorname{softplus}
        \bigl(F_{\lambda}(\bar{\mathbf{h}}_e)+\mathbf{b}_{\lambda}\bigr).
\end{aligned}
\end{equation}
The learned scalar $a$ is shared across key channels within a head,
and $\mathbf{b}_{\lambda}\in\mathbb{R}^{d_k}$ is a learned bias.
$F_{\lambda}:\mathbb{R}^{d_w}\rightarrow\mathbb{R}^{d_k}$ denotes
the head's output from two consecutive bias-free
linear layers, with no intervening activation.
Gate projection in Figure~\ref{fig:architecture}(c) comprises
these erase, write, and decay branches.

\noindent\textbf{Decay for image sequences.}
Spatial tokens within an observation share a source timestamp.
Applying temporal decay at every spatial token would compound
decay within one image, making the total decay depend on its token count.
We apply $\mathbf{D}_e$ once before processing the entry's tokens.
The tokenwise implementation uses log-decay vectors
$\mathbf{g}_{e,j}\in\mathbb{R}^{d_k}$:
\begin{equation}
    \mathbf{g}_{e,j} =
    \begin{cases}
        -\Delta_e\boldsymbol{\lambda}_e, & j=1, \\
        \mathbf{0}, & j>1.
    \end{cases}
\end{equation}
Thus, $\operatorname{diag}(\exp(\mathbf{g}_{e,1}))=\mathbf{D}_e$,
and subsequent tokens introduce no additional temporal decay.
The interval $\Delta_e$ is the difference between source-frame indices
of consecutive valid written entries, divided by the memory sampling
stride, with $\Delta_e=1$ at the first write.
Interaction entries use their paired observations' indices.
Decay therefore follows elapsed source time, while token updates
remain sequential and order-dependent. Applying decay before the
token updates also avoids temporally decaying information written
earlier within the same entry, as would occur if decay were
distributed across its tokens.
Appendix~\ref{app:decay_granularity} compares entry-level and
per-token decay on MIKASA-Robo.

\noindent\textbf{Scalar erasure.}
Gated DeltaNet-2 uses channel-wise erase and write gates
\cite{hatamizadeh2026gated}. We retain channel-wise writing but use
one erase scalar per token and head to address the numerical
instability observed with channel-wise erasure in our training runs.
After entry-level decay, each token applies the correction
\begin{equation}
\begin{aligned}
    \mathbf{r}_{e,j}
        &= \mathbf{w}_{e,j}\odot\mathbf{v}_{e,j}
           -\beta_{e,j}\mathbf{S}_{e,j-1}^{\top}\mathbf{k}_{e,j}, \\
    \mathbf{S}_{e,j}
        &= \mathbf{S}_{e,j-1}
           +\mathbf{k}_{e,j}\mathbf{r}_{e,j}^{\top},
\end{aligned}
\end{equation}
where $\mathbf{r}_{e,j}\in\mathbb{R}^{d_v}$ is the correction vector
and $\mathbf{S}_{e,0}=\mathbf{D}_e\mathbf{S}_{e-1}$.
For a unit-norm key $\mathbf{k}$ and erase gate $\beta\in(0,1)$,
the erase operator
$\mathbf{I}-\beta\mathbf{k}\mathbf{k}^{\top}$ has eigenvalue
$1-\beta$ along $\mathbf{k}$ and eigenvalue $1$ in orthogonal
directions. Its spectral norm is therefore at most one.
The channel-wise counterpart is generally asymmetric and does not
have this spectral-norm guarantee. Together with
$\|\mathbf{D}_e\|_2\le 1$, scalar erasure makes direct propagation
of the previous state non-expansive for fixed keys and gates.
This property concerns the propagated state; additive writes
can still increase its norm.
Appendix~\ref{app:erase_granularity} provides numerical diagnostics
comparing channel-wise and scalar erasure.

\noindent\textbf{State handling.}
Padded entries leave both the state and its source timestamp
unchanged. An empty history is represented by a zero state.
Retrieval reads the stored state without updating or decaying it.

\subsection{Storage and Retrieval Complexity}
\label{app:complexity}

\begin{figure*}[t]
    \centering
    \includegraphics[width=\textwidth]{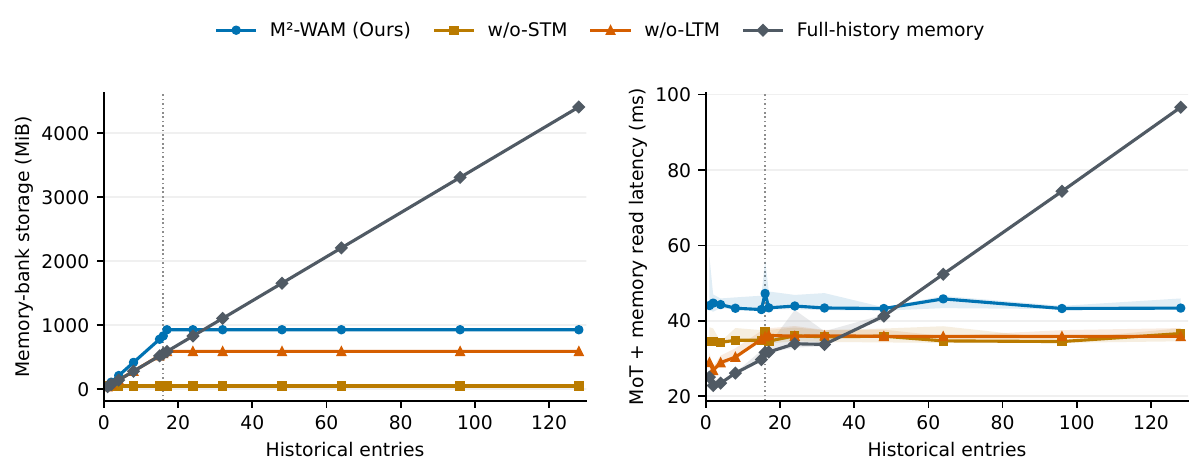}
    \caption{\textbf{Storage and retrieval scaling across memory designs.}
    Left: persistent memory-bank tensor storage.
    Right: MoT and memory-read latency per denoising iteration.
    Dotted lines mark FIFO capacity $N=16$; latency shading shows
    the range of medians across five measurement rounds. Full-history storage and read latency
    grow with history, while bounded FIFOs and fixed-size states
    yield storage plateaus and approximately stable read latency.}
    \label{fig:appendix_memory_complexity}
\end{figure*}

We analyze persistent memory storage and history-dependent computation
for a single sample. Model parameters and the backbone's processing
of current observations and actions are excluded from these
complexity expressions. Let $T$ denote
the number of accumulated historical entries, $J_{\mathrm v}$ the
number of spatial tokens per visual entry across views, and $R$
the number of action steps per executed prefix. We retain $N$,
$H$, and $P$ for FIFO capacity, prediction horizon, and tokens per
encoded prefix. We hold layer counts, head counts, and feature
dimensions fixed throughout the analysis.

\noindent\textbf{Storage and retrieval.}
Let $B_{\mathrm{s}}(n)$ denote the storage for $n$ paired visual
and interaction entries. Visual entries retain layer-wise KV tensors,
whereas interaction entries retain normalized executed prefixes,
giving $B_{\mathrm{s}}(n)=O(n(J_{\mathrm v}+R))$ with fixed
feature dimensions omitted. After prefix encoding, the readers attend
to $nJ_{\mathrm v}$ visual tokens and $nP$ interaction tokens.
For $H$ action queries, their history-dependent retrieval cost per
denoising iteration is
\[
    C_{\mathrm{s}}(n)=O\!\left(Hn(J_{\mathrm v}+P)\right).
\]
Each LTM head instead stores $d_kd_v$ scalars and reads its state
in $O(Hd_kd_v)$ operations. Summing over the fixed numbers of heads
and reader layers gives LTM storage $B_{\mathrm{l}}=O(1)$ and retrieval cost
$C_{\mathrm{l}}=O(H)$, both independent of $T$.
The interaction reader's computation is counted at each insertion
layer, although its parameters are shared.

\begin{table}[t]
    \centering
    \caption{\textbf{Persistent memory storage and retrieval complexity.}
    Retrieval costs are per denoising iteration; constant factors
    and fixed visual-anchor costs are omitted.
    Among these designs, only Full-history memory scales with
    accumulated history.}
    \label{tab:memory_complexity}
    \small
    \begin{tabular}{lll}
        \toprule
        Memory design & Storage & Retrieval \\
        \midrule
        Full-history memory & $B_{\mathrm{s}}(T)$ & $C_{\mathrm{s}}(T)$ \\
        w/o-LTM & $B_{\mathrm{s}}(N)$ & $C_{\mathrm{s}}(N)$ \\
        w/o-STM & $B_{\mathrm{l}}$ & $C_{\mathrm{l}}$ \\
        M\textsuperscript{2}-WAM
            & $B_{\mathrm{s}}(N)+B_{\mathrm{l}}$
            & $C_{\mathrm{s}}(N)+C_{\mathrm{l}}$ \\
        \bottomrule
    \end{tabular}
\end{table}

Table~\ref{tab:memory_complexity} compares the four designs once
the FIFO is full; before then, $N$ is replaced by $\min(T,N)$.
The expressions suppress constant factors, including differences
in the KV tensors retained by each implementation.
Full-history memory grows linearly with $T$ in both storage and
retrieval. The other designs have bounded costs for fixed $N$
and per-step input sizes.
Compared with w/o-LTM, M\textsuperscript{2}-WAM adds fixed-size
states to consolidate history beyond the FIFO. Compared with
w/o-STM, it retains recent detail at a bounded additional cost.
The w/o-STM variant writes entries directly into LTM and retains
the visual anchor without maintaining historical FIFOs.

\noindent\textbf{Preparation and updates.}
Interaction prefixes are encoded once per replanning step, before
denoising, with cost $O(n(R+P))$ for fixed encoder dimensions.
Their attention KV projections are recomputed at each reader
invocation. Retrieval costs recur at every denoising iteration.
Consolidation processes the evicted entry's $J_{\mathrm v}$ visual
tokens and $P$ encoded interaction tokens, with each per-head
state update costing $O(d_kd_v)$ per token.
For M\textsuperscript{2}-WAM, these preparation and update costs
are bounded with respect to $T$ because the number of retained
entries and the size of each evicted entry are bounded.
The bound applies to memory operations at each replanning step.
Cumulative computation still grows with the number of steps,
and training-time LTM reconstruction depends on the replay length.

\noindent\textbf{Empirical scaling.}
We measure the four implementations on a PPU-ZW810E with batch
size one, $N=16$, $H=32$, $R=16$, and $J_{\mathrm v}=98$, using
the same pretrained parameters for corresponding modules.
We profile system costs by cyclically replaying eight recorded
observations and executed prefixes with increasing timestamps,
varying $T$ up to 128. Storage is the sum of unique persistent
tensor allocations: shared allocations are counted once, and
model weights and temporary tensors are excluded.
Timing covers MoT attention and memory reads across all action
layers, including the backbone attention computation within MoT.
Instruction cross-attention, FFNs, prefix encoding, and memory
updates fall outside the timed regions.
Each condition uses five rounds of 50 measurements after 20
warmup iterations per round; plotted values are the median of
the round medians.

Figure~\ref{fig:appendix_memory_complexity} shows the predicted
scaling trends. Full-history memory exhibits linear storage growth and
increasing read latency, whereas M\textsuperscript{2}-WAM reaches
a storage plateau after the first FIFO eviction and maintains
roughly stable read latency as history grows. Removing either
STM or LTM reduces the measured overhead while omitting
FIFO-based recent context or associative consolidation, respectively.
The full hierarchy retains both at a bounded additional cost.
These instrumented module measurements support the scaling
comparison; end-to-end policy efficiency is evaluated separately
in Section~\ref{sec:inference_efficiency}.

\section{Experimental Details}
\label{app:experiments}

This section details the simulation benchmarks, real-world tasks,
and evaluation protocols, followed by the memory variant
configurations and retrieval interventions.

\subsection{Simulation Benchmarks}
\label{app:simulation}

We use benchmark-specific observation and control configurations,
summarized in Table~\ref{tab:simulation_inputs}. Each policy receives
RGB images, proprioception, and a task instruction. Shared training
settings, FIFO capacities, and action horizons are given in
Appendix~\ref{app:architecture}.

\begin{table}[t]
    \centering
    \caption{\textbf{Simulation observation and action configurations.}
    Image sizes are height $\times$ width in pixels after combining
    camera views. qpos denotes joint-position targets;
    action dimensions include gripper commands.}
    \label{tab:simulation_inputs}
    \small
    \begin{tabular}{@{}lccc@{}}
        \toprule
        Setting & MIKASA-Robo & RMBench & RoboTwin 2.0 \\
        \midrule
        Camera views
            & \begin{tabular}[c]{@{}c@{}}Top\\Wrist\end{tabular}
            & \begin{tabular}[c]{@{}c@{}}Head, front\\Left wrist\\Right wrist\end{tabular}
            & \begin{tabular}[c]{@{}c@{}}Main\\Left wrist\\Right wrist\end{tabular} \\
        \addlinespace
        Image size & $224\times448$ & $448\times448$ & $384\times320$ \\
        Action space & qpos & qpos & qpos \\
        Action mode & Absolute & Absolute & Absolute \\
        Action dim. & 8 & 14 & 14 \\
        \bottomrule
    \end{tabular}
\end{table}

\noindent\textbf{MIKASA-Robo.}
We jointly train on ShellGameTouch, InterceptMedium, and
RememberColor3/5/9 using the SimpleMemVLA demonstration dataset
\cite{yin2026simplememvla}, which provides 500 trajectories per
task, for 2,500 in total. Top-view and wrist-view RGB images
are resized to $224\times224$ and concatenated horizontally.
The policy predicts eight-dimensional absolute commands comprising
seven arm joint-position targets and one gripper command, using
the corresponding joint-position controller during evaluation.
Actions and the eight-dimensional proprioceptive state use
min--max normalization.

\noindent\textbf{RMBench.}
We use the nine-task suite in Table~\ref{tab:rmbench}, comprising
five $M(1)$ tasks and four $M(n)$ tasks. Under RMBench's task
memory-complexity definition, $M(1)$ requires retaining one
task-relevant past observation to disambiguate the current state,
whereas $M(n)$ requires retaining multiple such observations
for optimal decision-making~\cite{chen2026rmbench}.
The demonstration corpus contains 50 trajectories per task.
We jointly train one policy across the nine tasks using
episode-level instructions.
Observations include two static camera views and two wrist-camera
views, each resized to $224\times224$. The head and front views
form the upper row of a $2\times2$ image grid, with the left and
right wrist views below. The policy predicts 14-dimensional
absolute joint-position and gripper commands and receives a
14-dimensional proprioceptive state. Actions and proprioception
use z-score normalization. In Table~\ref{tab:rmbench}, ObsPick and BlocksRank
abbreviate Observe and Pick Up and Blocks Ranking, respectively.

\noindent\textbf{RoboTwin~2.0.}
We train a multi-task policy across all 50 tasks using a mixture
of 50 Clean and 500 Randomized demonstrations per task, totaling
2,500 and 25,000 trajectories, respectively. The same policy is
evaluated separately in the Clean and Randomized settings.
The main camera image occupies the upper $256\times320$ region
of the input, while two $128\times160$ wrist-camera images form
the lower row. The policy predicts 14-dimensional absolute
joint-position and gripper commands and receives a 14-dimensional
proprioceptive state. Both actions and proprioception use
z-score normalization.

\noindent\textbf{Evaluation.}
The main comparisons use 100 episodes per task, with the two
RoboTwin~2.0 settings evaluated separately. Memory is reset at
each episode boundary. Success is determined by the benchmark's
task-specific evaluator. For MIKASA-Robo, an episode succeeds
if its success condition is reached at any point during the
rollout. The retrieval interventions use the matched-seed protocol
in Appendix~\ref{app:intervention_protocol}.

\subsection{Real-World Tasks and Protocols}
\label{app:real_world}

\begin{figure*}[t]
    \centering
    \includegraphics[width=\textwidth]{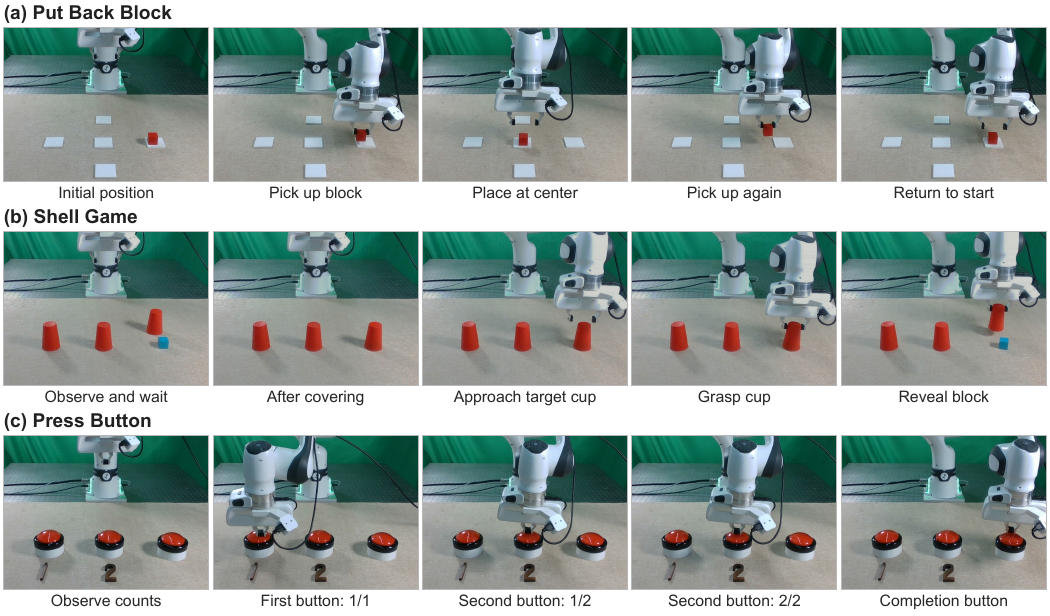}
    \caption{\textbf{Real-world task sequences.}
    Demonstration frames progress from left to right.
    \textbf{(a)} Put Back Block: move the block to the center, then return it
    to its original location.
    \textbf{(b)} Shell Game: use the initial scene to identify and lift
    the cup covering the block.
    \textbf{(c)} Press Button: press the first two buttons the indicated
    number of times, then press the completion button.
    Frames show task milestones at nonuniform intervals.}
    \label{fig:real_world_task_sequences}
\end{figure*}

We evaluate on Put Back Block, Shell Game, and Press Button using
the robot platform shown in Figure~\ref{fig:experimental_setup}.
Figure~\ref{fig:real_world_task_sequences} illustrates the task
procedures with representative demonstration frames.

\noindent\textbf{Platform and policy inputs.}
Our platform consists of a 7-DoF Franka Research~3 arm, a
main-view RealSense D435i camera, and a wrist-mounted RealSense
D405 camera. For M\textsuperscript{2}-WAM, each RGB image is
center-cropped to $640\times640$, resized to $224\times224$,
and concatenated horizontally with the other view to form a
$224\times448$ input. The eight-dimensional proprioceptive state
contains the end-effector position, quaternion orientation,
and gripper width. Actions comprise end-effector translation
increments, relative rotations represented by quaternions,
and a binary gripper command. Press Button uses a fixed closed gripper,
so its policy predicts only the seven pose-increment components.
We apply min--max normalization to actions and proprioception.

\begin{figure}[t]
    \centering
    \includegraphics[width=0.5\linewidth]{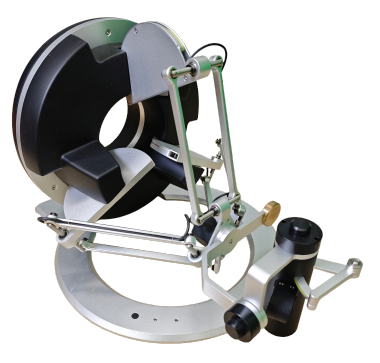}
    \caption{Omega.7 teleoperation device used to collect real-world
    demonstrations.}
    \label{fig:omega7_teleoperation}
\end{figure}

\noindent\textbf{Demonstrations and training.}
We collected 50 demonstration trajectories per task using an
Omega.7 teleoperation device, shown in
Figure~\ref{fig:omega7_teleoperation}, and trained a separate
M\textsuperscript{2}-WAM policy for each task for up to 20,000
optimization steps. Optimization and memory hyperparameters
are listed in Appendix~\ref{app:architecture}.

\noindent\textbf{Put Back Block.}
The robot moves a block to the center of the workspace and then
returns it to its original location. It must remember the initial
placement and track whether the intermediate move is complete.
Success requires completing both stages in order.

\noindent\textbf{Shell Game.}
Each trial begins with a five-frame observation phase, during
which the robot waits while receiving the initial scene observations.
The program then pauses while a human operator covers the block
with a cup. After the program resumes, the robot combines
its memory of the initial scene with the current observation
to identify and lift the cup covering the block.
A trial succeeds when the robot lifts the correct cup.

\noindent\textbf{Press Button.}
The scene contains three buttons, with a number displayed in
front of each of the first two. The robot tracks its completed
presses and must match both indicated counts before pressing
the third button to signal completion. A trial succeeds when
both counts are correct and the completion button is pressed last.

\noindent\textbf{Deployment and evaluation.}
Inference runs on a single NVIDIA RTX~3090 GPU, with control
commands issued at 15~Hz. M\textsuperscript{2}-WAM predicts
action chunks of length $H=32$ and executes the first 16 commands before
replanning from the latest observation. Both M\textsuperscript{2}-WAM
and Fast-WAM use precomputed text embeddings without loading
the text encoder during inference. We reset memory at the start
of each trial and evaluate each policy over 50 consecutive
trials per task. Trials exceeding 1,000 control steps are
counted as failures due to timeout.

\subsection{Memory Variant Configurations}
\label{app:variants}

We train the five variants in Table~\ref{tab:mikasa} and
Table~\ref{tab:rmbench} separately, using the same training data
and optimization budget as the full model. All retain the
permanent visual anchor, current-observation conditioning, and
benchmark-specific action horizon and memory sampling stride.
Each variant uses its memory configuration during both training
and evaluation.

\noindent\textbf{w/o-LTM.}
We remove both associative states and their writers and readers,
while retaining the visual and interaction FIFOs, each with
capacity $N$. MoT attention reads visual history, and interaction
cross-attention reads encoded executed prefixes.
Evicted entries are discarded without consolidation.

\noindent\textbf{w/o-STM.}
We remove both FIFOs while retaining the visual and interaction
associative states. Each sampled
observation is written directly into visual LTM before action
denoising, while its executed prefix is encoded and written into
interaction LTM after execution. The LTM readers remain,
while interaction STM cross-attention is removed. Apart from the
visual anchor, history is retained only in the associative states.

\noindent\textbf{Full-history memory.}
We retain all sampled visual KV entries and executed
prefixes from the episode, without eviction or LTM consolidation.
Action queries read the accumulated visual history through MoT
attention and the encoded prefixes through interaction
cross-attention. Storage and
retrieval costs consequently grow with episode length, as analyzed
in Appendix~\ref{app:complexity}.

\noindent\textbf{w/o modality separation.}
We retain the STM--LTM hierarchy but use a shared memory pathway
for both modalities. The prefix encoder produces $P$ interaction
tokens, which receive learned within-prefix position embeddings
and layer-specific projections into visual KV space. The projected
interaction keys and values are concatenated with their visual
counterparts from the paired observation to form one FIFO entry.
Entries are committed after prefix execution and evicted as a unit.
On eviction, visual tokens followed by interaction tokens update
the same per-layer associative state. Retrieval uses the visual
STM and LTM pathways, without separate interaction readers.
The FIFO retains a capacity of $N$, and
relative-age embeddings are shared by the tokens within each entry.

\noindent\textbf{w/o-interaction memory.}
We remove the executed-prefix encoder, interaction FIFO and LTM,
and their readers. The visual anchor, visual FIFO, and visual LTM
remain unchanged.

\subsection{Memory Retrieval Intervention Protocol}
\label{app:intervention_protocol}

We assess the effect of memory retrieval using a fixed checkpoint
on ShellGameTouch, InterceptMedium, and RememberColor3/5/9.
For each task, all four retrieval conditions use the same
30 environment seeds, yielding 30 rollouts per condition.
Figure~\ref{fig:memory_retrieval_intervention} reports their
success rates.

\noindent\textbf{Retrieval conditions.}
\emph{Full retrieval} enables both STM and LTM retrieval.
\emph{STM read off} disables retrieval from the visual and
interaction FIFOs while retaining both LTM readers.
\emph{LTM read off} disables both associative-state readouts
while retaining retrieval from the two FIFOs.
\emph{Both reads off} combines these interventions.
All four conditions retain current-observation conditioning
and access to the permanent initial visual anchor.

The selected retrieval condition remains active throughout each
episode, across all denoising iterations and relevant layers.
FIFO storage, eviction, and LTM updates proceed normally;
only retrieval is masked.

\noindent\textbf{Paired evaluation.}
We reset memory at each episode boundary and use the same task
success criteria as in Appendix~\ref{app:simulation}.
Matching seeds controls the initial task conditions, while
closed-loop actions and subsequent observations may diverge
across retrieval conditions.

\section{Additional Memory Design Analyses}
\label{app:additional_memory_design}

We examine sensitivity to FIFO capacity, memory use without a
permanent visual anchor, and the effects of erase and decay
granularity.

\subsection{Sensitivity to FIFO Length}
\label{app:fifo_length}

We examine sensitivity to visual FIFO capacity at inference
using a fixed checkpoint trained with $N=4$.
We evaluate $N\in\{0,1,2,8,16\}$ over 100 episodes per task
on the five MIKASA-Robo tasks, retaining visual LTM and the
permanent initial anchor throughout.
Table~\ref{tab:fifo_capacity} reports task success rates.

\begin{table}[htbp]
    \centering
    \caption{\textbf{Success rates (\%) across visual FIFO capacities.}
    Shell and Intercept denote ShellGameTouch and InterceptMedium;
    RC3/5/9 follow Table~\ref{tab:mikasa}.
    $\dagger$ denotes an earlier evaluation of the same checkpoint,
    included for reference and not rerun in this sweep.}
    \label{tab:fifo_capacity}
    \small
    \setlength{\tabcolsep}{3pt}
    \begin{tabular}{@{}c|ccccc|c@{}}
        \toprule
        $N$ & Shell & Intercept & RC3 & RC5 & RC9 & Avg. \\
        \midrule
        0 & 97 & 51 & 70 & 50 & 35 & 60.6 \\
        1 & 94 & 63 & 72 & 51 & 38 & 63.6 \\
        2 & 93 & 80 & 74 & 51 & 39 & 67.4 \\
        4$^{\dagger}$ & 98 & 91 & 81 & 63 & 61 & 78.8 \\
        8 & 91 & 73 & 74 & 51 & 39 & 65.6 \\
        16 & 91 & 74 & 74 & 51 & 39 & 65.8 \\
        \bottomrule
    \end{tabular}
\end{table}

Sensitivity to FIFO capacity varies across tasks.
InterceptMedium benefits substantially from recent visual context,
with success increasing from 51\% at $N=0$ to 80\% at $N=2$.
ShellGameTouch remains above 90\% across all evaluated capacities,
while RememberColor3/5/9 show smaller changes across the newly
evaluated settings. These differences suggest that tasks vary in
their reliance on the amount of recent history retained for direct
retrieval. FIFO capacity can therefore be selected according to
task requirements and the available memory budget.

\begin{figure}[t]
    \centering
    \includegraphics[width=\linewidth]{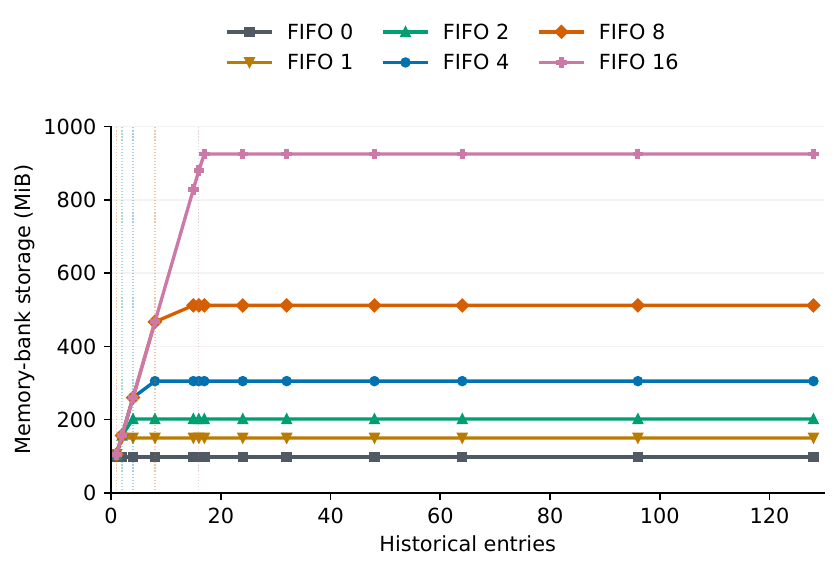}
    \caption{\textbf{Memory-bank storage under different FIFO capacities.}
    Storage is measured in MiB as historical entries accumulate.
    Larger capacities yield higher storage plateaus.}
    \label{fig:fifo_storage}
\end{figure}

Figure~\ref{fig:fifo_storage} shows that larger FIFOs increase
memory-bank storage, while each fixed capacity reaches a plateau
as history accumulates. Once the FIFO is full, evicted entries
are consolidated into fixed-size associative states.
Increasing capacity therefore raises the storage budget for
recent observations while preserving bounded storage over
longer histories.

\subsection{Effect of the Visual Anchor}
\label{app:visual_anchor}

The permanent visual anchor gives the policy direct access to the
initial observation. To examine memory use without this permanent access, we train
and evaluate a variant without the anchor on the five MIKASA-Robo
tasks, while retaining the STM--LTM hierarchy.
Table~\ref{tab:anchor_off} compares this variant with the full
M\textsuperscript{2}-WAM model from Table~\ref{tab:mikasa}.

\begin{table}[t]
    \centering
    \caption{\textbf{Effect of the visual anchor on MIKASA-Robo success rates (\%).}
    The w/o-anchor variant omits the anchor during training and evaluation.
    Task abbreviations follow Table~\ref{tab:fifo_capacity}.}
    \label{tab:anchor_off}
    \small
    \setlength{\tabcolsep}{3pt}
    \begin{tabular}{@{}l|ccccc|c@{}}
        \toprule
        Method & Shell & Intercept & RC3 & RC5 & RC9 & Avg. \\
        \midrule
        \oursrow{}
        M\textsuperscript{2}-WAM & \textbf{98} & \textbf{91} & 81 & \textbf{63} & \textbf{61} & \textbf{78.8} \\
        w/o-anchor & 93 & 77 & \textbf{83} & 61 & \textbf{61} & 75.0 \\
        \bottomrule
    \end{tabular}
\end{table}

Removing the anchor reduces average success only modestly, from
78.8\% to 75.0\%. This 3.8-point drop is substantially smaller
than the 15.4-point drop for w/o-LTM in Table~\ref{tab:mikasa},
which achieves 63.4\% with the anchor retained.
The policy thus retains most of its performance without a
permanent visual anchor, supporting LTM's role in
preserving earlier information beyond the STM window.

\subsection{Effect of Erase Granularity}
\label{app:erase_granularity}

\begin{figure*}[t]
    \centering
    \includegraphics[width=\textwidth]{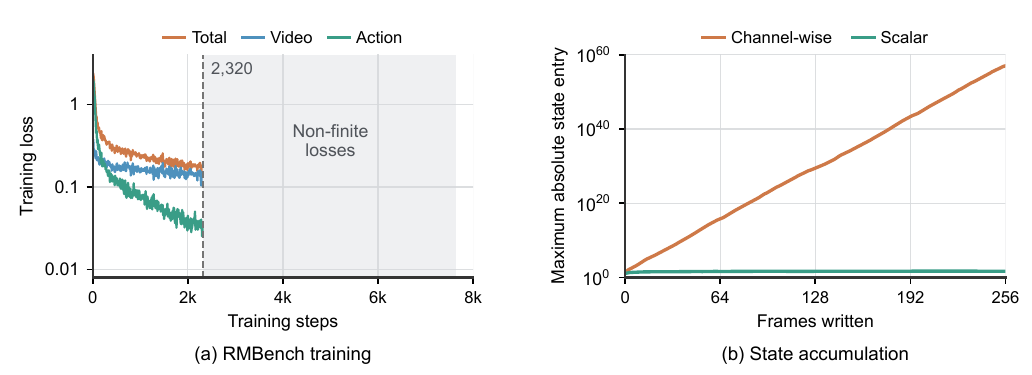}
    \caption{\textbf{Training instability and the effect of erase granularity.}
    \textbf{(a)} RMBench training losses with channel-wise erasure;
    shading marks the interval with non-finite losses.
    \textbf{(b)} Maximum absolute state entries during cyclic replay
    of the same input sequence under the two erase rules.
    Channel-wise erasure produces sustained state growth,
    whereas scalar erasure suppresses this amplification.}
    \label{fig:erase_stability}
\end{figure*}

During training, we observe numerical instability when using
channel-wise erasure in visual memory, motivating our use of
one erase scalar per token and head.
In an RMBench training run, the total, video, and action losses
first become non-finite at logged step 2,320 and remain non-finite
through the final record at step 7,640, as shown in
Figure~\ref{fig:erase_stability}(a).

To understand the failure mechanism, consider the erase operator
for a normalized key $\mathbf{k}_{e,j}$ and a channel-wise gate
$\boldsymbol{\beta}^{\mathrm{ch}}_{e,j}\in(0,1)^{d_k}$:
\begin{equation}
    \mathbf{E}^{\mathrm{ch}}_{e,j}
    = \mathbf{I}-\mathbf{k}_{e,j}
      (\boldsymbol{\beta}^{\mathrm{ch}}_{e,j}
       \odot\mathbf{k}_{e,j})^{\top}.
\end{equation}
Unlike the scalar erase operator in Appendix~\ref{app:gated_delta},
this matrix is generally asymmetric and can amplify the state.
This distinction matters for image sequences: temporal decay is
applied once per observation, while successive spatial-token
updates can accumulate amplification within that observation.

To diagnose this training instability, Figure~\ref{fig:erase_stability}(b)
compares the two erase rules using a trained memory writer.
The same input sequence is replayed cyclically without resetting
the state.
We replace each channel-wise erase gate with its
per-token, per-head channel mean, keeping keys, values, decay,
and write gates unchanged.
With channel-wise erasure, the maximum absolute state entry grows
steadily to approximately $10^{57}$ after 256 frames.
Scalar erasure instead keeps the state magnitude at a much smaller
scale, with the maximum absolute entry remaining below 56 even
when replay is extended to 1,024 frames.
This comparison shows that scalar erasure suppresses state
amplification in the replay, supporting its use to address the
numerical instability observed during training while retaining
channel-wise writing.

\subsection{Effect of Decay Granularity}
\label{app:decay_granularity}

We compare applying temporal decay at every spatial token with
our entry-level decay, which applies it once per visual entry.
The per-token variant is evaluated after 30,000 training steps
on the five MIKASA-Robo tasks.
Table~\ref{tab:decay_granularity} compares its success rates
with the entry-level results from Table~\ref{tab:mikasa}.

\begin{table}[t]
    \centering
    \caption{\textbf{Effect of decay granularity on MIKASA-Robo success rates (\%).}
    Entry-level decay is used in M\textsuperscript{2}-WAM.
    Task abbreviations follow Table~\ref{tab:fifo_capacity}.}
    \label{tab:decay_granularity}
    \small
    \setlength{\tabcolsep}{3pt}
    \begin{tabular}{@{}l|ccccc|c@{}}
        \toprule
        Decay & Shell & Intercept & RC3 & RC5 & RC9 & Avg. \\
        \midrule
        \oursrow{}
        Entry-level & \textbf{98} & \textbf{91} & \textbf{81} & \textbf{63} & \textbf{61} & \textbf{78.8} \\
        Per-token & 97 & 89 & 62 & 60 & 57 & 73.0 \\
        \bottomrule
    \end{tabular}
\end{table}

Entry-level decay achieves higher success on all five tasks,
improving average success from 73.0\% to 78.8\%.
The largest gain is 19 percentage points on RememberColor3,
compared with one and two points on ShellGameTouch and
InterceptMedium, respectively.
Spatial tokens within one observation share a timestamp, yet
per-token decay repeatedly attenuates the state as these tokens
are processed. Our entry-level rule instead applies decay once
according to the elapsed source time, independently of the number
of spatial tokens, as detailed in Appendix~\ref{app:gated_delta}.
These results support separating temporal decay from within-frame
spatial-token updates when adapting gated delta updates to image sequences.

\begin{figure*}[t]
    \centering
    \includegraphics[width=\textwidth]{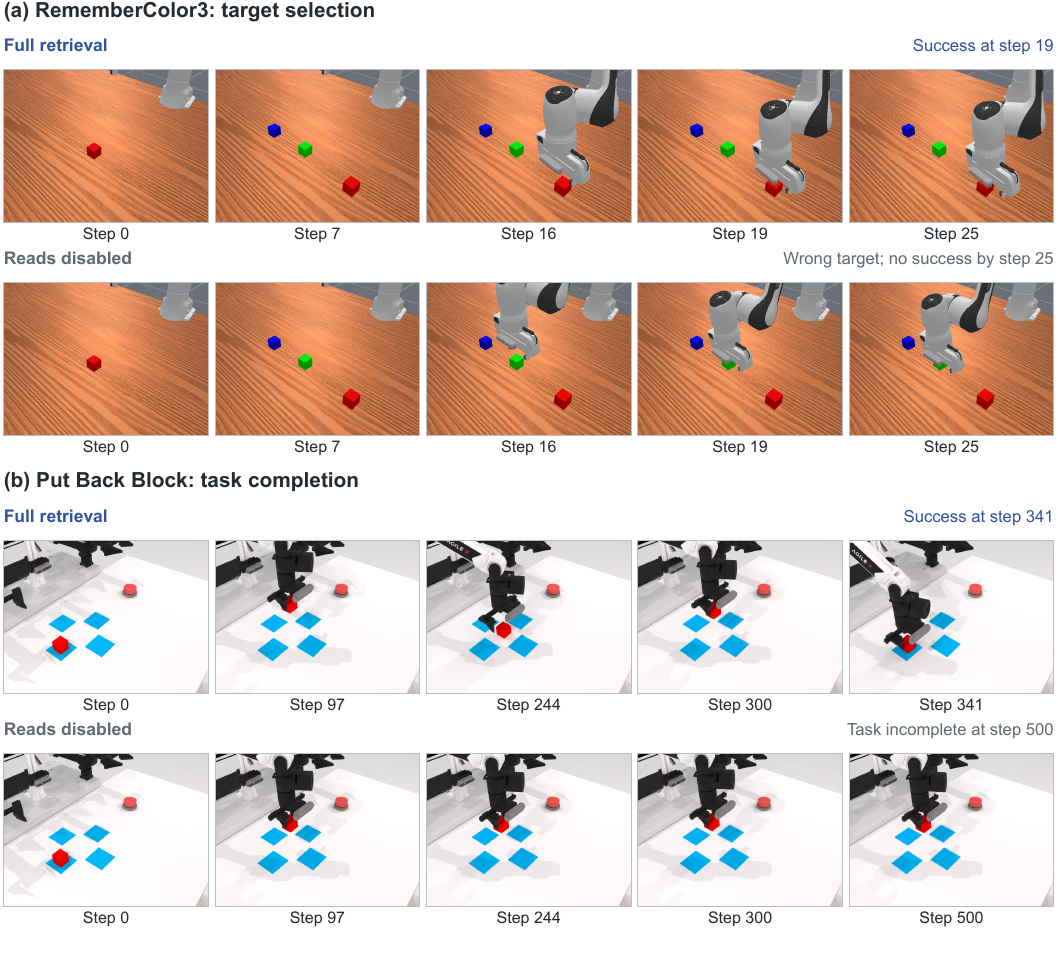}
    \caption{\textbf{Qualitative effects of disabling memory retrieval.}
    Each pair uses the same checkpoint and environment seed.
    \textbf{(a)} In RememberColor3, the robot reaches the red target
    with full retrieval,
    whereas disabling visual memory and anchor reads leads the robot
    toward a green distractor.
    \textbf{(b)} In Put Back Block, disabling visual and interaction memory
    reads, including the anchor, leaves the robot holding the block
    without completing the task.}
    \label{fig:appendix_read_cases}
\end{figure*}

\section{Qualitative Analysis and Failure Cases}
\label{app:qualitative}

This section examines how memory retrieval affects decisions,
compares real-world trajectories with Fast-WAM, and analyzes
simulation and real-world failures.

\begin{figure*}[t]
    \centering
    \includegraphics[width=\textwidth]{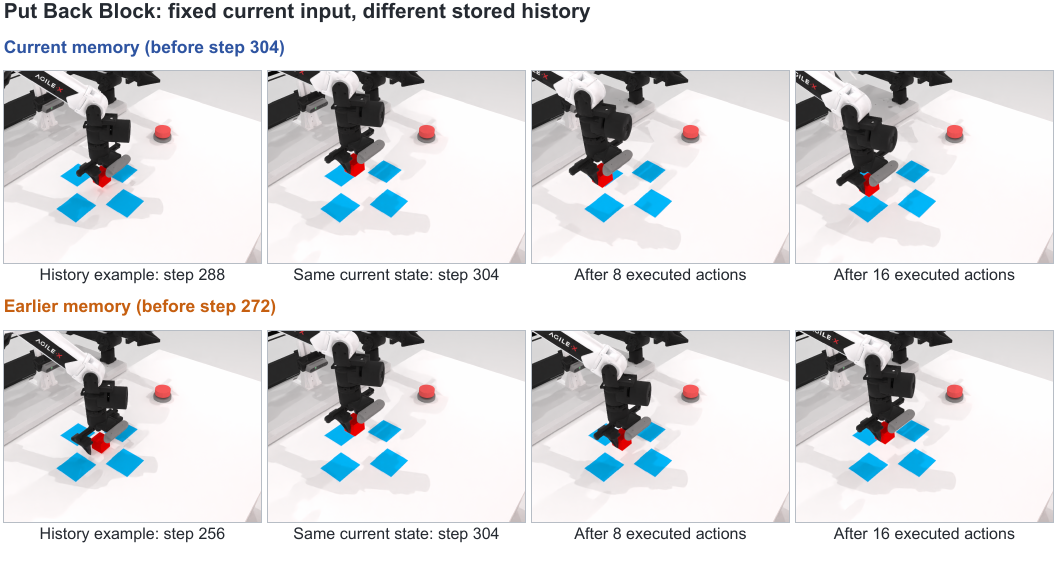}
    \caption{\textbf{Effect of stored history with current inputs fixed.}
    In Put Back Block, the two rows use memory snapshots from before
    steps 304 and 272, respectively, while sharing the current inputs
    and simulation state at step~304. The first column shows the
    most recent observation stored in each snapshot. The remaining
    columns show the shared starting state and the states after
    8 and 16 executed actions. With all readers enabled, changing
    the stored history produces different subsequent movements.}
    \label{fig:appendix_history_replacement}
\end{figure*}

\subsection{Memory Retrieval Intervention Cases}
\label{app:intervention_cases}

Figure~\ref{fig:appendix_read_cases} compares rollouts with memory
retrieval enabled or disabled, using the same checkpoint and
environment seed for each pair. Current observations remain
available, and memory writes proceed normally. The RememberColor3
case uses visual memory, while the Put Back Block case uses both
visual and interaction memories. In both cases, memory is queried
by action features.

In RememberColor3, retrieval enables the policy to reach the
initially indicated red target at step~19. Without retrieval,
it approaches the green distractor and does not complete the task
within 25 steps.
In Put Back Block, the policy succeeds at step~341 with retrieval;
without it, the robot grasps the block but holds it above the
workspace until the 500-step limit. These behaviors illustrate
how retrieved history supports target selection and task progression.

\noindent\textbf{Changing history with the current input fixed.}
In Figure~\ref{fig:appendix_history_replacement}, we replace the
Put Back Block memory bank at step~304 with a snapshot from before
step~272 of the same episode. Current images, proprioception,
instruction, warm-start actions, and sampling seed remain fixed,
with all readers enabled. Executing both predicted prefixes from
the same reconstructed simulation state yields right end-effector trajectories
that differ by up to 8.10\,cm over 16 actions. Thus, stored history
changes decisions even with identical current inputs.

\begin{figure*}[t]
    \centering
    \includegraphics[width=\textwidth]{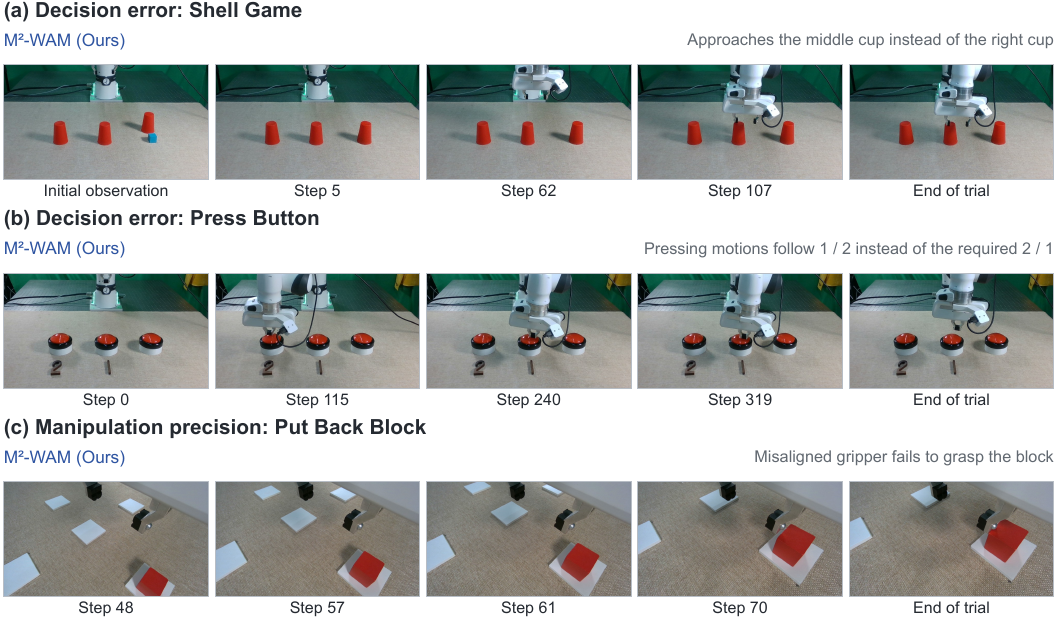}
    \caption{\textbf{Decision and manipulation-precision failures of M\textsuperscript{2}-WAM.}
    \textbf{(a)} Shell Game: the robot approaches the middle cup
    although the initial cue identifies the right cup.
    \textbf{(b)} Press Button: the robot makes one downward pressing motion
    at the first button and two at the second, reversing the required counts.
    \textbf{(c)} Put Back Block: gripper misalignment prevents a
    stable grasp. Panels (a,b) use the main view; panel (c) uses
    the wrist view to show the contact geometry. Labels identify
    recorded steps, initial observations, or trial endpoints.}
    \label{fig:appendix_real_failures}
\end{figure*}

\begin{figure*}[t]
    \centering
    \includegraphics[width=\textwidth]{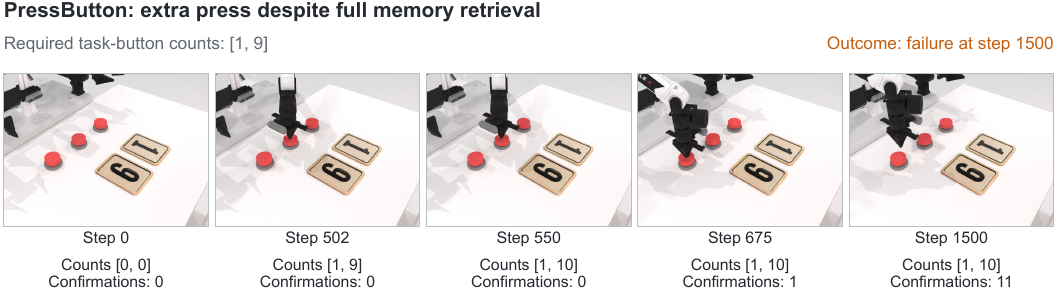}
    \caption{\textbf{PressButton failure with memory retrieval enabled.}
    The policy reaches the required task-button counts $[1,9]$ but
    continues to $[1,10]$ before repeatedly pressing the confirmation
    button. The task remains unsuccessful. Labels report control
    steps, task-button counts, and confirmation counts from the
    environment's event log.}
    \label{fig:appendix_press_failure}
\end{figure*}

\begin{figure*}[t]
    \centering
    \includegraphics[width=\textwidth]{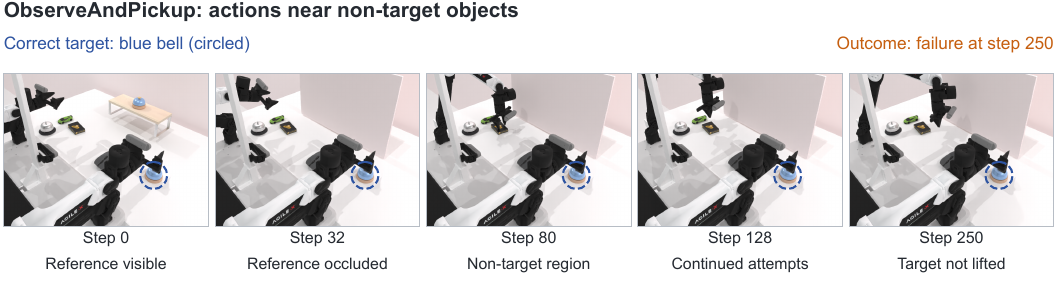}
    \caption{\textbf{ObserveAndPickup failure with memory retrieval enabled.}
    The initial reference specifies the matching blue bell on the
    tabletop, marked by a dashed blue circle in each frame.
    After the reference is occluded, the robot operates
    near non-target objects and fails to lift the target within 250 steps.
    Frames are captured from a front-right display camera, with
    control steps labeled.}
    \label{fig:appendix_oap_failure}
\end{figure*}

\subsection{Real-World Trajectory Comparisons}
\label{app:real_world_cases}

\begin{figure*}[p]
    \centering
    \includegraphics[width=\textwidth]{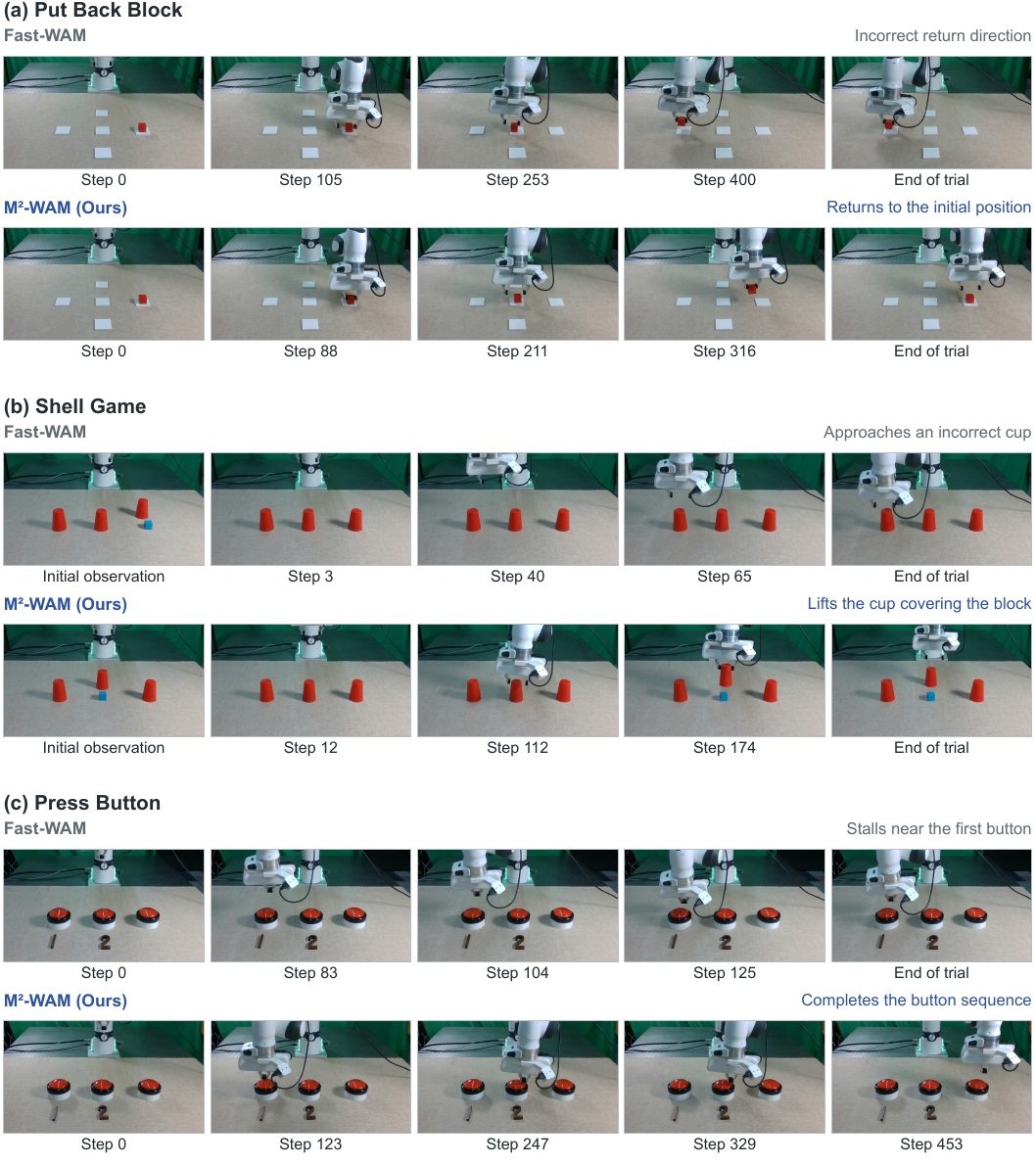}
    \caption{\textbf{Representative real-world trajectories of Fast-WAM and M\textsuperscript{2}-WAM.}
    \textbf{(a)} Put Back Block: M\textsuperscript{2}-WAM returns the block
    to its initial position, while Fast-WAM moves toward a different location.
    \textbf{(b)} Shell Game: M\textsuperscript{2}-WAM reveals the hidden
    block, while Fast-WAM approaches an incorrect cup.
    \textbf{(c)} Press Button: M\textsuperscript{2}-WAM completes the
    button sequence, while Fast-WAM stalls near the first button.
    Rows show separate trials. Labels indicate recorded steps,
    initial observations, or trial endpoints.}
    \label{fig:appendix_real_comparisons}
\end{figure*}

Figure~\ref{fig:appendix_real_comparisons} complements the aggregate
results in Section~\ref{sec:main_results} with representative trials.
In Put Back Block, both policies grasp the block and move it to
the center. Fast-WAM then carries it toward the opposite side,
whereas M\textsuperscript{2}-WAM returns it to its initial position.
In Shell Game, M\textsuperscript{2}-WAM lifts the cup covering the
block, while Fast-WAM approaches an incorrect cup. These cases
illustrate how historical context supports destination and target
selection when current observations alone are insufficient.

Press Button also requires progression through repeated actions.
M\textsuperscript{2}-WAM presses the first task button, repeats the
motion at the second, and proceeds to the completion button.
Fast-WAM instead remains near the first button without completing
the sequence, illustrating a failure in task progression.

\begin{figure*}[t]
    \centering
    \includegraphics[width=\textwidth]{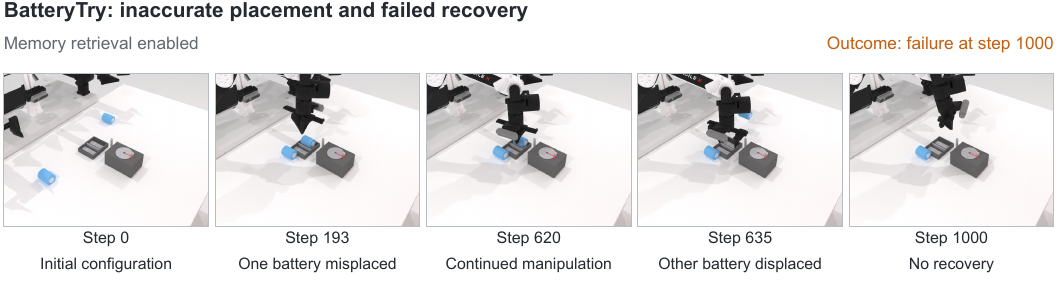}
    \caption{\textbf{BatteryTry failure with memory retrieval enabled.}
    One battery remains beside its slot, while subsequent manipulation
    dislodges the other from an initially aligned position.
    The robot fails to recover before the 1,000-step limit.
    Frames show a close-up display view at the labeled control steps.}
    \label{fig:appendix_battery_failure}
\end{figure*}

\subsection{Failure Cases}
\label{app:failure_cases}

\noindent\textbf{PressButton.}
With both memories enabled in Figure~\ref{fig:appendix_press_failure},
the robot reaches the required counts $[1,9]$ at step~502 but
presses the second task button again, reaching $[1,10]$ at step~550.
It then presses the confirmation button 11 times without success
by the 1,500-step limit. This failure to transition at the required
count may reflect limitations in the Fast-WAM backbone's
generalization and character recognition, as discussed in
Section~\ref{sec:main_results}.

\noindent\textbf{ObserveAndPickup.}
In Figure~\ref{fig:appendix_oap_failure}, both memories are enabled,
but the robot fails to pick up the blue bell matching the initial
reference. After the reference is occluded, the left arm repeatedly
operates near non-target objects, while the right arm remains
nearly stationary. The target remains unmoved at the 250-step limit.

Across our experiments, the policy performs well on layouts
resembling the training data but fails more often on unfamiliar
arrangements, even for the same target. This sensitivity highlights
limited generalization, consistent with Section~\ref{sec:main_results}.
We hypothesize that the policy relies too heavily on the current
scene, associating familiar layouts with trajectories rather than
consistently using the initial reference to select the target.

\noindent\textbf{BatteryTry.}
The task requires placing two batteries in their slots with the
correct orientations. With both memories enabled in
Figure~\ref{fig:appendix_battery_failure}, one battery remains beside
its slot, while the other is initially aligned but subsequently
knocked off the tabletop. Neither placement is recovered before
the 1,000-step limit. This case highlights difficulties in
preserving intermediate progress and recovering from placement errors.

\noindent\textbf{Real-world decision errors.}
Figure~\ref{fig:appendix_real_failures}(a,b) shows target-selection
and action-sequencing errors with M\textsuperscript{2}-WAM.
In Shell Game, the robot approaches the middle cup despite the
initial cue identifying the right cup. In Press Button, it makes
one downward pressing motion at the first button and two at the
second, reversing the displayed requirements. These errors reflect
incorrect use of initial target cues or numerical instructions.

\noindent\textbf{Real-world manipulation precision.}
In Figure~\ref{fig:appendix_real_failures}(c), the wrist view reveals
gripper misalignment in Put Back Block. The robot fails to grasp
the block, which remains on its original pad. Such errors may also
reflect limitations in low-level tracking accuracy, highlighting
the need to coordinate high-level action predictions with
low-level control.

\section{Limitations and Future Work}
\label{app:limitation}

\noindent\textbf{Selective memory writing and retention.}
Apart from the initial anchor, dense visual KV storage is allocated
by recency, although older observations may contain fine-grained details needed
later. The FIFO sensitivity in Appendix~\ref{app:fifo_length} also
motivates selectively retaining older observations or spatial
regions within a fixed budget. Replacing the fixed sampling stride
with writes guided by scene changes, task progress, or anticipated
decision relevance could further reduce redundant updates.

\noindent\textbf{Adaptive memory retrieval.}
Retrieval follows a predefined schedule across layers and
denoising iterations, although the need for history varies with
task stage and current observations. Adaptive readers could choose
when to retrieve and which modalities or memory levels to consult.
Skipping unnecessary reads or reusing context across denoising
iterations could further reduce computation.

\noindent\textbf{Memory use and generalization.}
The failures in Appendix~\ref{app:failure_cases} suggest that memory
use does not generalize consistently to unfamiliar layouts.
Broader layout variation and examples with similar current
observations but different historical cues and required actions
could encourage more reliable use of history. Finer visual or
contact feedback could complement memory conditioning to improve
grasp alignment and manipulation precision.

\let\addcontentsline\supplementOriginalAddcontentsline